%% file: main.tex
\documentclass{article}

\usepackage{microtype}
\usepackage{graphicx}
\usepackage{booktabs} 
\usepackage{enumitem}
\usepackage{multirow}
\usepackage[dvipsnames]{xcolor}
\usepackage{cuted}
\usepackage{capt-of}
\usepackage{braket}
\usepackage{float}
\usepackage{afterpage}
\usepackage{listings}
\usepackage{stfloats}
\usepackage{subcaption}
\usepackage{xcolor} 
\definecolor{lightbrown}{rgb}{0.95, 0.9, 0.8}

\usepackage{hyperref}

\usepackage[accepted]{icml2024}

\usepackage{amsmath}
\usepackage{amssymb}
\usepackage{mathtools}
\usepackage{amsthm}

\usepackage[capitalize,noabbrev]{cleveref}

\theoremstyle{plain}

\theoremstyle{definition}

\theoremstyle{remark}

\newcommand{\task}{\tau}

\usepackage[textsize=tiny]{todonotes}
\definecolor{bluegray}{rgb}{0.4, 0.6, 0.8}

\icmltitlerunning{BBSEA: An Exploration of Brain-Body Synchronization for Embodied Agents}

\begin{document}

\twocolumn[
\icmltitle{BBSEA: An Exploration of Brain-Body Synchronization for Embodied Agents}



\icmlsetsymbol{equal}{*}

\begin{icmlauthorlist}
\icmlauthor{Sizhe Yang}{equal,yyy,sch}
\icmlauthor{Qian Luo}{equal,yyy}
\icmlauthor{Anumpam Pani}{yyy}
\icmlauthor{Yanchao Yang}{yyy}

\end{icmlauthorlist}

\icmlaffiliation{yyy}{University of Hong Kong}
\icmlaffiliation{sch}{University of Electronic Science and Technology of China}

\icmlcorrespondingauthor{Sizhe Yang}{3599699144yang@gmail.com}
\icmlcorrespondingauthor{Qian Luo}{luoqian1@connect.hku.hk}

\icmlkeywords{LLM, Embodied AI}

\vskip 0.3in
]


\printAffiliationsAndNotice{\icmlEqualContribution} 

\input{sec/0_abstract}

\input{sec/1_introduction}

\input{sec/2_related_work}

\input{sec/3_method}
\input{sec/4_experiments}
\input{sec/5_conclusion}

\bibliography{main}
\bibliographystyle{icml2024}

\newpage
\input{sec/X_suppl}

\end{document}

%% file: sec/0_abstract.tex
\begin{abstract}
Embodied agents capable of complex physical skills can improve productivity, elevate life quality, and reshape human-machine collaboration. We aim at autonomous training of embodied agents for various tasks involving mainly large foundation models. It is believed that these models could act as a brain for embodied agents; however, existing methods heavily rely on humans for task proposal and scene customization, limiting the learning autonomy, training efficiency, and generalization of the learned policies. In contrast, we introduce a brain-body synchronization ({\it BBSEA}) scheme to promote embodied learning in unknown environments without human involvement. The proposed combines the wisdom of foundation models (``brain'') with the physical capabilities of embodied agents (``body''). Specifically, it leverages the ``brain'' to propose learnable physical tasks and success metrics, enabling the ``body'' to automatically acquire various skills by continuously interacting with the scene. 
We carry out an exploration of the proposed autonomous learning scheme in a table-top setting, and we demonstrate that the proposed synchronization can generate diverse tasks and develop multi-task policies with promising adaptability to new tasks and configurations. 
We will release our data, code, and trained models to facilitate future studies in building autonomously learning agents with large foundation models in more complex scenarios.
More visualizations are available at \href{https://bbsea-embodied-ai.github.io}{https://bbsea-embodied-ai.github.io}
\end{abstract}

%% file: sec/1_introduction.tex
\section{Introduction}
\label{sec:intro}

\begin{figure*}[!ht]
    \centering
    \includegraphics[width=0.95\linewidth]{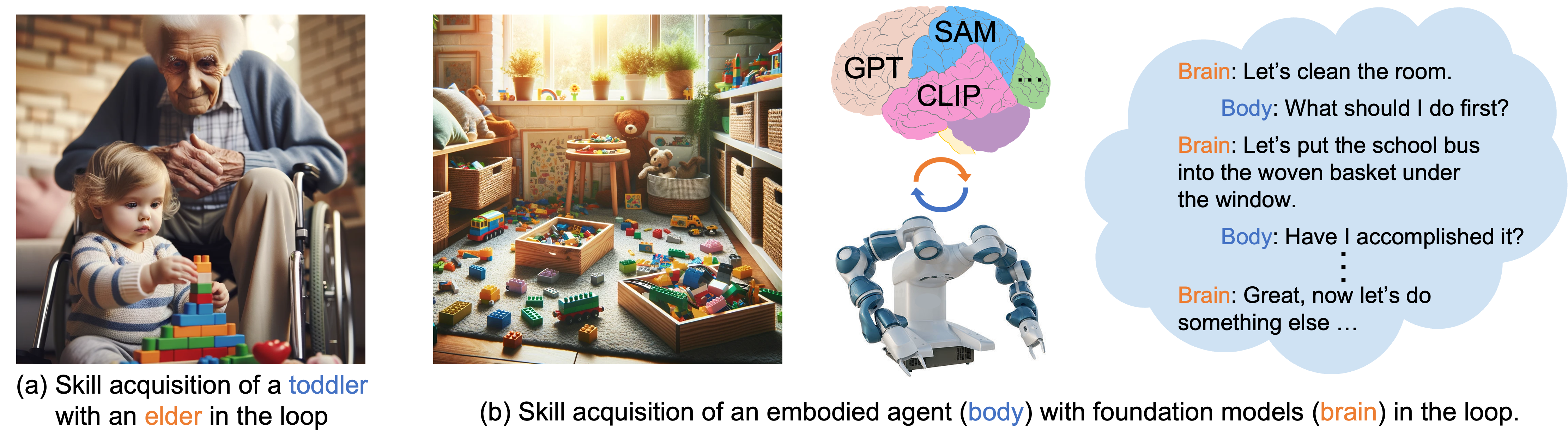}
    \vspace{-5mm}
    \caption{
    Left: An experienced elder instructs a toddler on playing with colorful blocks (image credit to GPT-4V). Right: Using the proposed brain-body synchronization scheme, foundation models ({\color{orange} brain}) teach an embodied agent ({\color{bluegray} body}) a variety of physical interaction skills. To accomplish this, the {\color{orange} brain} needs to propose interaction tasks that are compatible with the scene and the {\color{bluegray} body}'s physical constraints, as well as define measurable success metrics for the suggested tasks. The {\color{bluegray} body} then synchronizes with the {\color{orange} brain} through trial and error, acquiring interaction skills solely based on feedback from the {\color{orange} brain}.}
    \vspace{-2mm}
    \label{fig:teaser}
\end{figure*}

With age, one may lose the capacity to act, to see, and even to hear;
however, it also comes with wisdom such that one can think smartly and view clearly through the mind’s eye.
Thus, it is important to learn from senior people, whose rich experience and structural reasoning can significantly facilitate the development of the younger ones (Fig.~\ref{fig:teaser} (a)).
Nowadays, given the advancement of large foundation models (LFMs), a similar complementarity emerges for the development of embodied agents.
On one hand, we have models like GPT \citep{radford2018improving,radford2019language,brown2020language,OpenAI2023GPT4TR} that can perform reasoning in the language space as well as question and answer with visual input.
However, these models are not grounded in the 3D environment (embodiment) and are incapable of physical interactions.
On the other, we have embodied agents that can interact with the scene but lack an understanding of the physical world to accomplish semantically meaningful tasks.

In this work,
we treat the ensemble of the foundation models (e.g., GPTs) as a wise elderly equipped with the common sense of the physical world (the brain), and seek the possibilities of training an embodied agent (the body) from scratch with the disembodied structural knowledge of the brain towards intelligent physical interactions.
More explicitly,
we consider the scenarios in which the foundation models, together with an embodied agent (of a basic factory setting), are put into an environment without any prior knowledge of the specific scene.
Our goal is to develop a pipeline that enables the autonomous acquisition of various physical interaction skills for the embodied agent with as little human involvement as possible. 

We propose that the key to our goal lies in threefold.
First, the {\it brain} has to propose interaction tasks or skills for the {\it body} to learn, which have to be compatible with the scene and achievable given the physical constraints of the agent.
Second, to make the proposed tasks learnable by the {\it body}, the {\it brain} should also define the tasks by specifying metrics that help determine whether the task is successfully executed or not.
Lastly, the {\it body} has to acquire the skills for accomplishing the proposed tasks through efficient interaction (trial and error) with solely the feedback (e.g., whether the metrics are satisfied) from the {\it brain}.
Existing works leveraging foundation models for training embodied agents either rely on humans to propose tasks \citep{ha2023scalingup} or synthesize scenes to suit the tasks proposed by LFMs \citep{wang2023gensim},
which imposes a bottleneck on the learning autonomy and is subject to potential generalization issues. 
In contrast, our approach is uniquely positioned to minimize human input in task proposal and scene customization, aiming to enhance autonomous learning ability and generalization in various settings and providing a foundation for future research in more complex environments.

To achieve autonomous learning of embodied agents in unknown scenes,
we develop a framework that first comes up with task proposals compatible with the scene and the agent's physical limitations.
Specifically, the scene is parsed with a few sensing modules, e.g., SAM \citep{SAM} and CLIP \citep{radford2021learning}, which extract rich semantic and physical information about the scene components.
A scene graph is then constructed holding this information and serving as an easy interface for an LFM to comprehend the environment and propose diverse and achievable interaction tasks.
Meanwhile, a task completion metric(s) is instantiated by prompting an LFM to provide feedback on whether a proposed task is accomplished based on the scene graphs (before and after action).
These two components form a verifiable definition of a task, i.e., its description and measurement of success, and also render the task learnable by an embodied agent.
Consequently, an agent performs brain-guided exploration with action primitives for obtaining successful execution trajectories which are further distilled into a language-conditioned policy for enhancing and expanding the agent's physical interaction capabilities.
Therefore, the name {\it brain-body synchronization} for the fact that an embodied agent can align with the physical understanding of a collection of foundation models via continuous interaction and feedback.

The proposed synchronization diagram helps improve the efficiency of learning interactive tasks and the explainability of the learned policies, e.g., the structural knowledge in foundation models can help reduce the exploration cost while the modular instructions can help understand the behavior of the embodied agents.
Moreover, it enhances the generalization capability of embodied agents (learned interactive policies) to new tasks, e.g., by reusing common skills summarized in the language guidance as well as efficiently acquiring novel skills.
Furthermore, language-driven human-robot cooperation can be automatically achieved with the minimum involvement of human agents in the training loop.
Experiments show that our system can automatically propose diverse and feasible tasks on par with humans and distill a multi-task policy in a continuous manner.
The learned policy also shows graceful zero-shot capabilities and can be effectively fine-tuned for novel tasks, evidencing adaptability in diverse physical interactions. 
Even though our study is mainly performed in a table-top scenario, which is commonly used in the literature, our exploration presents useful insights and a solid foundation for scaling up to more complex environments.

In summary, our contributions are:
\begin{itemize}[leftmargin=*]
    \vspace{-3mm}
    \item A framework that integrates foundation models with embodied agents for autonomously learning physical interaction tasks in unknown environments, creating an effective {\it brain-body synchronization} and a stepping stone towards more complex real-world scenarios.
    \vspace{-2mm}
    \item A task proposal module for efficient scene comprehension and automatically proposing {\it scene-compatible tasks}. Moreover, it establishes {\it metrics for task completion}, aiding the embodied agent in skill acquisition with minimum human involvement.
    \vspace{-2mm}
    \item An extensive validation of the proposed synchronization by continuously learning an effective language-conditioned policy in both {\it zero-shot} and {\it few-shot} settings, demonstrating reasonable adaptability to novel tasks and configurations.
\end{itemize}

%% file: sec/2_related_work.tex
\begin{figure*}[!ht]
  \centering
  \includegraphics[width=0.95\linewidth]{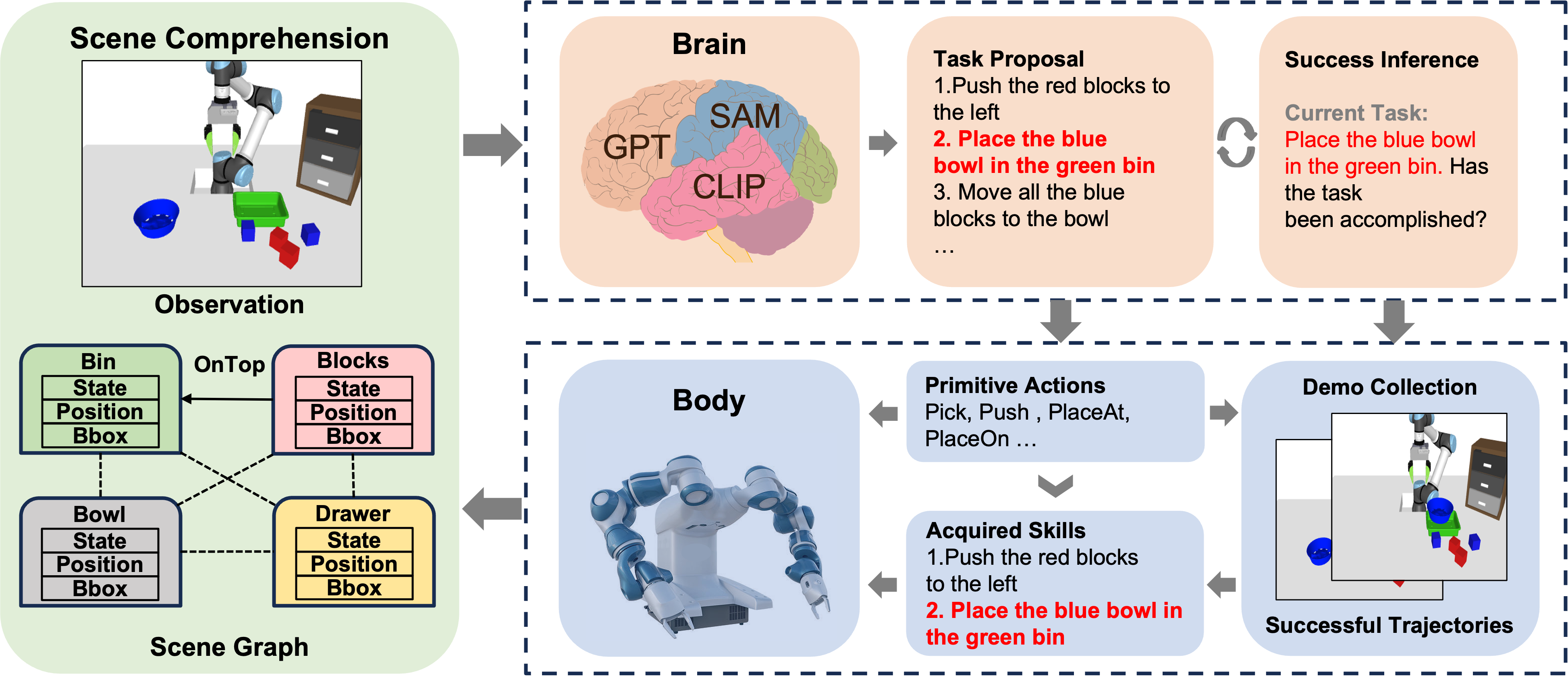}
  \vspace{-1mm}
  \caption{
  An overview of the proposed brain-body synchronization. 
  The scene comprehension module constructs and passes a scene graph of the current environment to an LFM ({\color{orange} brain}). 
  The {\color{orange} brain} then proposes interaction tasks compatible with the scene and the physical limitations of the {\color{bluegray} body}, which acquires the interaction skills via trial and error with solely the feedback from the {\color{orange} brain}.}
  \vspace{-1mm}
  \label{fig:bbsea-overview}
\end{figure*}

\section{Related Work}
\label{sec:related}

\paragraph{Embodied AI with Large Foundation Models.}
The advent of Large Foundation Models (LFMs), including Language Models (LLMs) and Vision-Language Models (VLMs), enables significant advancements in Embodied AI research, making progress across various domains such as manipulation \citep{zeng2022socratic,huang2022inner,liang2023code,lin2023text2motion,driess2023palm}, navigation \citep{khandelwal2022simple,shah2023lm,huang2023visual}, and locomotion \citep{tang2023saytap,yu2023language}. 
In addition, LLMs have been instrumental in transforming complex language instructions into robotic affordance, empowering tailored and flexible task execution \citep{ahn2022can,huang2022language,song2023llm,wu2023tidybot}. 
Furthermore, advances in code policy generation \citep{liang2023code,singh2023progprompt,huang2023voxposer} highlight LFMs' adaptability to varied environments, diverse robotic functionalities, and a spectrum of task requirements. 
Recent studies \citep{brohan2023rt,huang2023instruct2act,szot2023large} also reveal LFMs' role in shaping low-level policies for complex visual tasks in different scenarios.
Unlike the aforementioned research, our framework leverages the common sense knowledge from LFMs to acquire a rich set of skills in unknown environments.
Building on the emergent capabilities of LFMs as tools for embodied learning \citep{wang2023gensim,fang2022active}, reward design \citep{yu2023language,xie2023text2reward,ma2023eureka}, and completion inference \citep{liu2023reflect,di2023towards}, we develop a fully automated pipeline to learn policies with minimum human intervention.

\vspace{-2mm}
\paragraph{Policy Distillation in Robotic Manipulation.}
Robot learning has achieved great success in mastering manipulation skills via reinforcement learning \citep{gu2017deep,rajeswaran2017learning,yarats2021image,Hansen2022tdmpc} or imitation learning \citep{zhan2020framework,lynch21language,jang21bc-z,mees22what}. 
Recent research also leverages VLMs to train language-conditioned multi-task policies \citep{shridhar2022cliport,jiang2022vima,shridhar2022peract,rvt2023,luo2023grounding}. 
However, in these works, both the design of tasks and the supervision needed for policy learning require a substantial amount of human effort. 
LFMs help relieve humans of this labor to some extent. Huy et al. \citep{ha2023scalingup} scale up the data collection for manipulation skills by leveraging an LLM, then distilling the robot experiences into a visuo-linguistic-motor policy that infers control sequences. 
However, the diversity of skills is still constrained by the quantity of tasks available, which are defined by human experts. 
In this paper, we propose an innovative framework that automates diverse skill acquisition for robotic manipulation. 
Our method uses LFMs as a central {\it brain} to guide the robot {\it body}, enabling embodied agents to acquire various manipulation tasks efficiently, bypassing the need for human-provided task descriptions, success metrics, or demonstrations.

%% file: sec/3_method.tex
\begin{figure*}[!t]
  \centering
  \begin{subfigure}{0.295\linewidth}
    \includegraphics[width=0.95\textwidth]{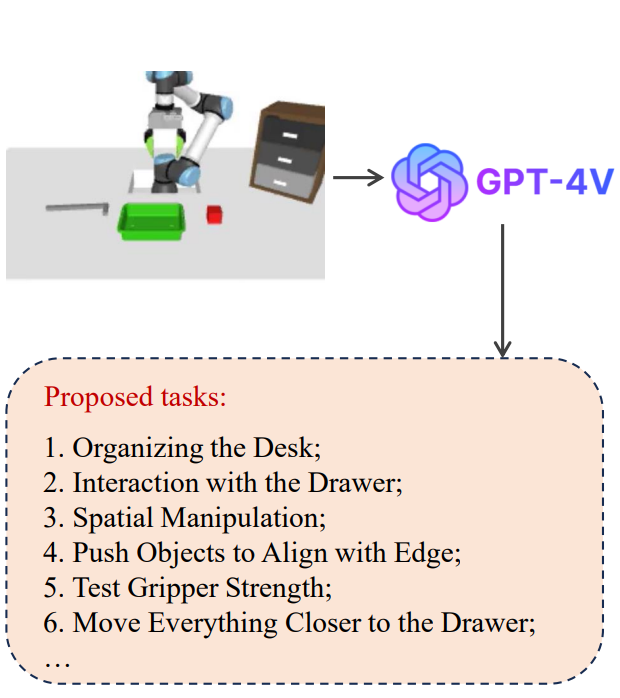}
    \caption{Tasks proposed directly with GPT-4V.}
    \label{fig:task-proposal-gpt4v}
  \end{subfigure}
  \hfill
  \begin{subfigure}{0.665\linewidth}
    \includegraphics[width=0.95\textwidth]{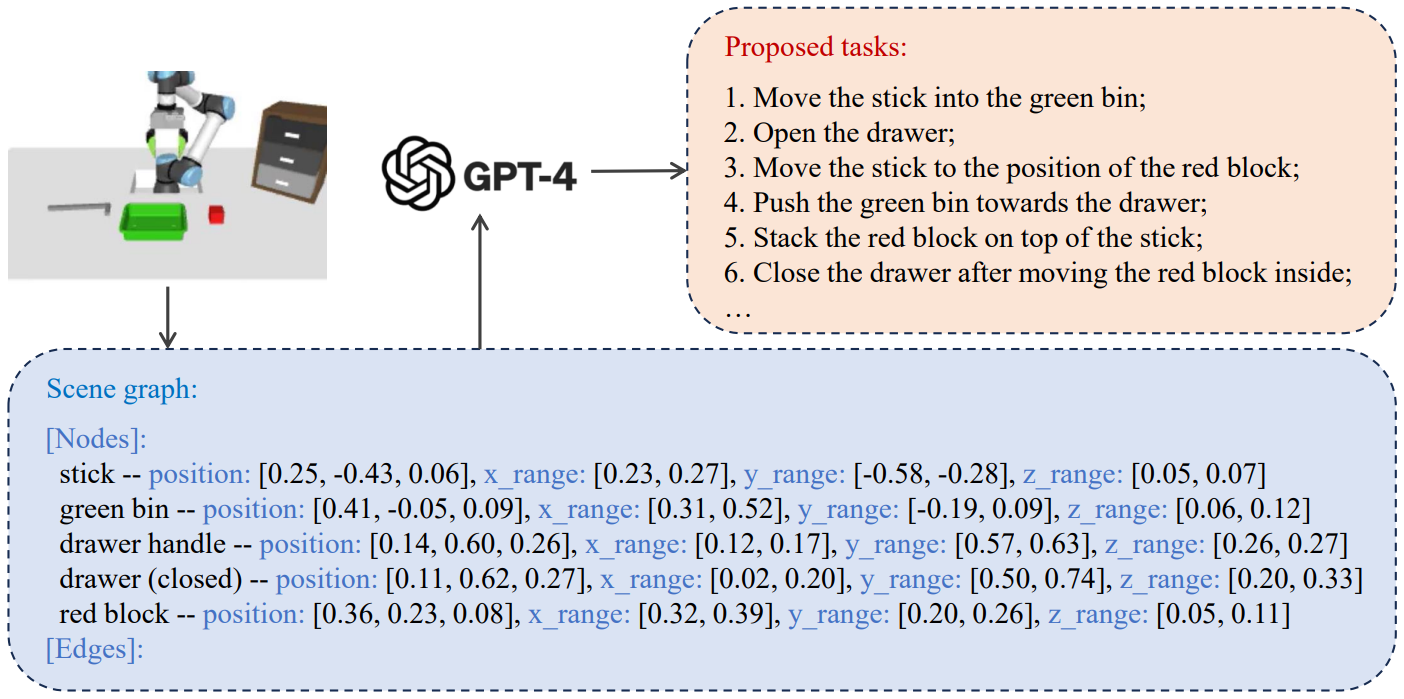}
    \caption{Tasks proposed with our method.}
    \label{fig:task-proposal-ours}
  \end{subfigure}
  \vspace{-2mm}
  \caption{Comparison between proposed tasks through GPT-4V (GPT4+Vision) and the task proposer in our pipeline. Ours can propose more context-relevant and feasible tasks for the agent to learn, leveraging easily digestible scene information in the graph.}
  \vspace{-3mm}
  \label{fig:task-proposal-ours-vs-gpt4v}
\end{figure*}

\section{Brain-Body Synchronization}
\label{sec:method}

Our goal is to convert common-sense knowledge in (disembodied) Large Foundation Models into manipulation policies supporting language-based instructions. 
Specifically, we aim to automate the training process for embodied agents in various manipulation tasks with minimum human intervention through the help of LFMs. 
Our proposed framework consists of two components: the {\it brain} -- foundation models and a perception module, and the {\it body} -- robot arms that interact with the environment based on textual instructions. 
The {\it brain} leverages a scene graph to understand the environment, proposes tasks, and determines task completion. 
And the {\it body} learns a policy for the tasks proposed by the {\it brain}.
In summary, our pipeline comprises three parts: 1) scene-compatible task proposal, 2) task completion inference, and 3) task-conditioned policy learning. An overview is shown in Fig.~\ref{fig:bbsea-overview}.

\subsection{Scene-Compatible Task Proposal}
\label{sec:scene-compatible-task-proposal}

Automatic task proposal and engagement in physical interactions are crucial for embodied agents to acquire diverse manipulation skills and adapt to new environments \citep{haywood2021life,haibach2023motor}. 
A {\it task} represents a set of reachable states from the current scene configuration, considering the agent's physical limitations (e.g., a wheeled robot can not climb the wall to install a light bulb). 
Next, we describe how to use an LFM to continuously suggest compatible manipulation tasks within an agent's physical constraints, initiating the skill acquisition process.

\subsubsection{Scene Comprehension with Robust Sensing}
\label{sec:scene-understanding}

First, it is important to have an accurate perception of scene entities' physical states. 
Although large models show promising physical reasoning capabilities, directly querying GPT-4V \citep{yang2023dawn} may result in unreasonable tasks due to imprecise and ungrounded understanding \citep{yang2023set} (Fig.~\ref{fig:task-proposal-ours-vs-gpt4v}). 
Thus, we resort to robust sensing modules to extract meaningful scene information and enable the LFM to produce scene-compatible tasks.

We assume two RGBD cameras for sensing: one horizontal view observing the robot and one top-down view for a holistic scene perspective. 
Using RGBD images, we first employ an object detector to identify objects \citep{Jocher_YOLOv5_by_Ultralytics_2020}, outputting 2D bounding boxes and names. 
Then we prompt SAM \citep{SAM} with detected bounding boxes to derive segmentation masks, refining spatial occupancy.
This enables extracting point clouds (3D position and bounding boxes) for each object, given camera parameters and depth.

In addition to ``what'' and ``where'' information, 
an LFM should know the objects' states (in text) to propose reasonable tasks. 
For each object, we predefine a list of semantically meaningful states depending on its category (e.g., a drawer can be $\{``open'', ``closed''\}$). 
Please refer to Appendix~\ref{sec:appendix-implementation-details} for a full list of the considered states.
We then apply CLIP encoders \citep{radford2021learning} on object segments and textual state descriptions to detect objects' states using cosine similarities. 
With the necessary information (object name, state, position, and bounding box) collected in text and numbers, we then leverage a scene graph to inform an LFM to propose executable tasks.

\vspace{-3mm}
\paragraph{Scene Graph Construction}
We organize perceived object information using a scene graph $G=(V,E)$ as input to the LFM, where $V$ represents object nodes, $v \in V = \{obj.name, state, position, bbox\}$, and $E$ is the set of edges. 
An edge $e \in E$ represents a spatial relationship ($r$) between objects $v_i$ and $v_j$ as $e=(v_i, r, v_j)$, and is determined using point clouds. 
If two nodes lack a meaningful spatial relationship, no edge is established. For more details, see Appendix~\ref{sec:appendix-implementation-details}.

\begin{figure*}[!ht]
  \centering
  \includegraphics[width=1.0\linewidth]{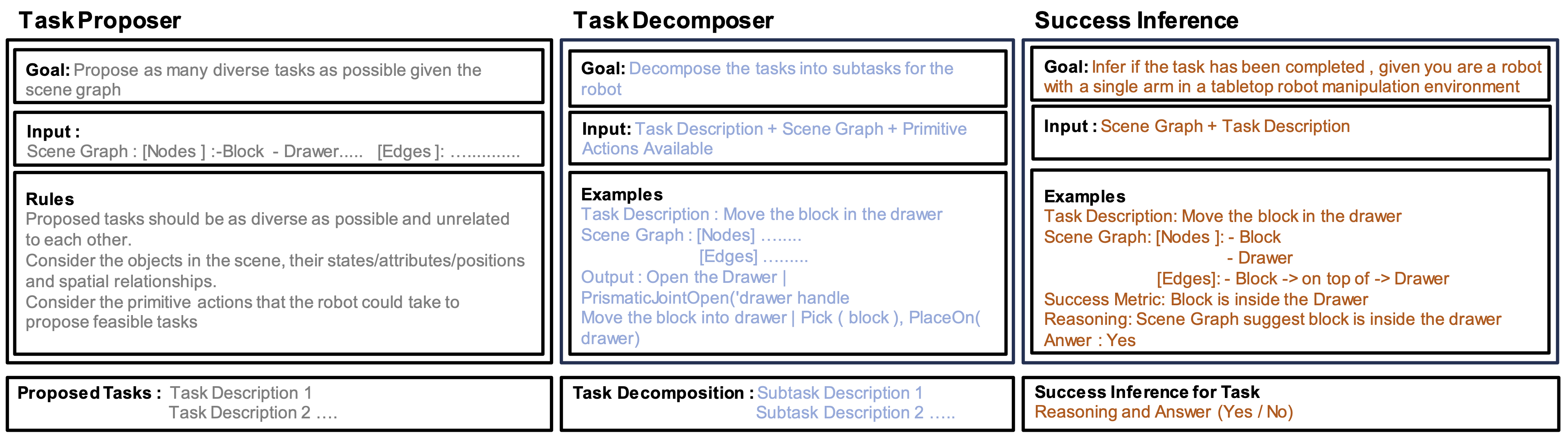}
  \vspace{-6mm}
  \caption{An overview of the prompts used for the Task Proposer (left), Task Decomposer (middle), and the Success Inference (right) modules. 
  These prompts ensure effective collection and completion inference of diverse and feasible tasks. 
  Please note that all the prompts are fixed without tailoring to a specific task.}
  \vspace{-2mm}
  \label{fig:prompt-eng}
\end{figure*}

\subsubsection{Prompting LLMs for Task Proposal}
\label{sec:task-proposal}

With the comprehensive information contained in the scene graph, we leverage LLMs' commonsense knowledge of the physical world to propose multiple interaction tasks to drive the learning.
More explicitly, an LLM takes in the scene graph $G$ and outputs a set of task descriptions $T(G)=\{\task_i\}$, which should comply with the following:
\begin{itemize}[leftmargin=*]
    \vspace{-3mm}
    \item The tasks are based on the objects in the scene, considering their states and spatial relationships. This ensures that the tasks are grounded and permitted by the scene.
    \vspace{-2mm}
    \item The tasks should also be aware of the agent's physical capabilities in terms of primitive actions. This ensures that the tasks are practically achievable.
    \vspace{-2mm}
    \item With the above, the LLM still needs to propose a wide range of tasks, ensuring diversity in the interactions and skills to be acquired. This is crucial for training a versatile and adaptable embodied agent.
\end{itemize}
To ensure these properties, 
we develop a prompting method shown in Fig.~\ref{fig:prompt-eng} (left).
The prompt consists of a clearly defined goal (what to output), along with a few (fixed) in-context examples or rules that the LLM could refer to while formulating the response, as well as the output format.
The devised prompt guarantees that 
the tasks proposed are feasible, diverse, and tailored to the specific environment and capabilities of the robot.
We perform a study of the diversity of the proposed manipulation tasks in Sec.~\ref{sec:diversity-analysis}.

Furthermore, to facilitate task-conditioned policy training, the LLM is asked to propose tasks that can be decomposed into sequences of predefined (factory) actions. This has two benefits. First, it prevents physically ambiguous tasks, like ``Organizing the Desk,'' which could imply numerous goal states. Second, explicit action items using primitives allow easy assessment of task success. Examples of proposed tasks are in Fig.~\ref{fig:task-proposal-ours}.

Hereto, we introduce an automatic mechanism using robust sensing modules and LLMs to propose scene-compatible, achievable, and definitive manipulation tasks. However, textual tasks need a measure of physical state to determine completion. We address this with the inference module introduced in the next section.

\subsection{Task Completion Inference}
\label{sec:task-completion-inference}

For acquiring interaction skills, 
the foundation models ({\it brain}) must propose diverse manipulation tasks and provide feedback on their completion. 
Assessing task completion is crucial for training embodied agents ({\it body}) to derive action policies. 
Thus, we utilize GPT-4's reasoning capability to automatically generate success criteria and determine task accomplishment.

Specifically, when an agent performs a sequence of actions, an updated scene graph $G'$ is constructed, reflecting object states and relationships. 
The before-action scene graph $G$ and after-action scene graph $G'$ are encapsulated in a prompt, providing GPT-4 with the necessary information to infer task completion status. An example prompt is shown in Fig.~\ref{fig:prompt-eng} (right), applicable to other tasks by changing the task description.

It is worth noting that GPT-4 can generalize beyond in-context examples. For instance, Fig.~\ref{fig:prompt-eng} shows decisions based on direction, but in practice, GPT-4 can also calculate object positions to assess task accomplishment. This ensures that an embodied agent receives meaningful instructions and effective feedback during learning.

\subsection{Task-Conditioned Policy Learning}
\label{sec:task-conditioned-policy-learning}

Given the proposed tasks and completion metrics,
training an embodied agent to efficiently achieve proposed tasks fulfills the synchronization between the foundation models ({\it brain}) and the agent ({\it body}). One approach is using reinforcement learning \citep{sutton2018reinforcement} with LLM-generated task completion status as the reward. However, this is challenging due to the large exploration space and sparse rewards. Instead, we use robustly executable primitive actions to collect successful task demonstrations and train a task-conditioned policy network via behavior cloning \citep{pomerleau1988alvinn}.

\vspace{-3mm}
\paragraph{Task Decomposition with Primitive Actions}
To collect demonstrations, 
we first prompt GPT-4 to decompose generated tasks into step-by-step instructions using primitive actions like ``Pick,'' ``PlaceOn,'' and ``Push,'' which cover most daily physical interaction tasks \citep{wu2023tidybot}. See Appendix~\ref{sec::appendix-implementation-details-primitive-actions} for the full list of primitive actions.
The task decomposition prompt is shown in Fig.~\ref{fig:prompt-eng} (middle).
\vspace{-3mm}
\paragraph{Demonstration Collection}
We employ a waypoint sampler and motion planner to execute primitive actions proposed by the LLM. 
The waypoint sampler takes object point clouds and GPT-4-generated parameters as input to sample potential waypoints, while the motion planner determines robot arm configurations to achieve these waypoints. Actions are executed sequentially and assessed using the completion inference mechanism. Successful task trajectories are added to the demonstration pool and collected for various tasks and scenes.
Note that the demonstration collection mechanism is not the task-conditioned policy to be learned. Its trial-and-error nature leads to less robust task performance, and fixed action primitives without trainable parameters limit learning and adaptation from interaction experiences, as described in the following.
\vspace{-3mm}
\paragraph{Task-Conditioned Policy} 
With successful trajectories, 
an embodied agent can continuously train a language-conditioned action policy to acquire diverse manipulation skills. 
The policy takes scene observations and task descriptions as input, outputting actions that achieve the desired goal state. Distilling physical experiences into a task-conditioned policy is crucial for an evolving automatic skill acquisition framework.

The conditional policy is learned using behavior cloning by training the network proposed in \citep{ha2023scalingup}. The 10-dimensional action space policy network allocates three dimensions for position, six for rotation (represented by elements in the upper two rows of the rotation matrix $\{ r_{i,j} \mid i \in \{0,1\}, j \in \{0,1,2\} \}$), and one for the gripper command. By learning a stream of tasks suggested by the scene-compatible proposal module, the embodied agent, equipped with the evolving conditional policy, gradually acquires skills to perform complex interaction tasks.

To this point, 
we present an autonomous method for training embodied agents to continuously acquire manipulation skills by transforming foundation models' high-level physical understanding into low-level motor commands, i.e., brain-body synchronization. 
We also note that online reinforcement learning can perform synchronization within a lifelong training framework. 
However, by reducing exploration overhead using behavior cloning and enabling continuous learning with rehearsal, we establish a minimal yet functionally equivalent pipeline to investigate essential modules' roles. 
Next, we evaluate the pipeline's effectiveness in acquisition efficiency and policy generalization.

%% file: sec/4_experiments.tex
\begin{figure*}[!t]
  \centering
  \begin{subfigure}{0.32\linewidth}
    \includegraphics[width=0.95\textwidth,height=2in]{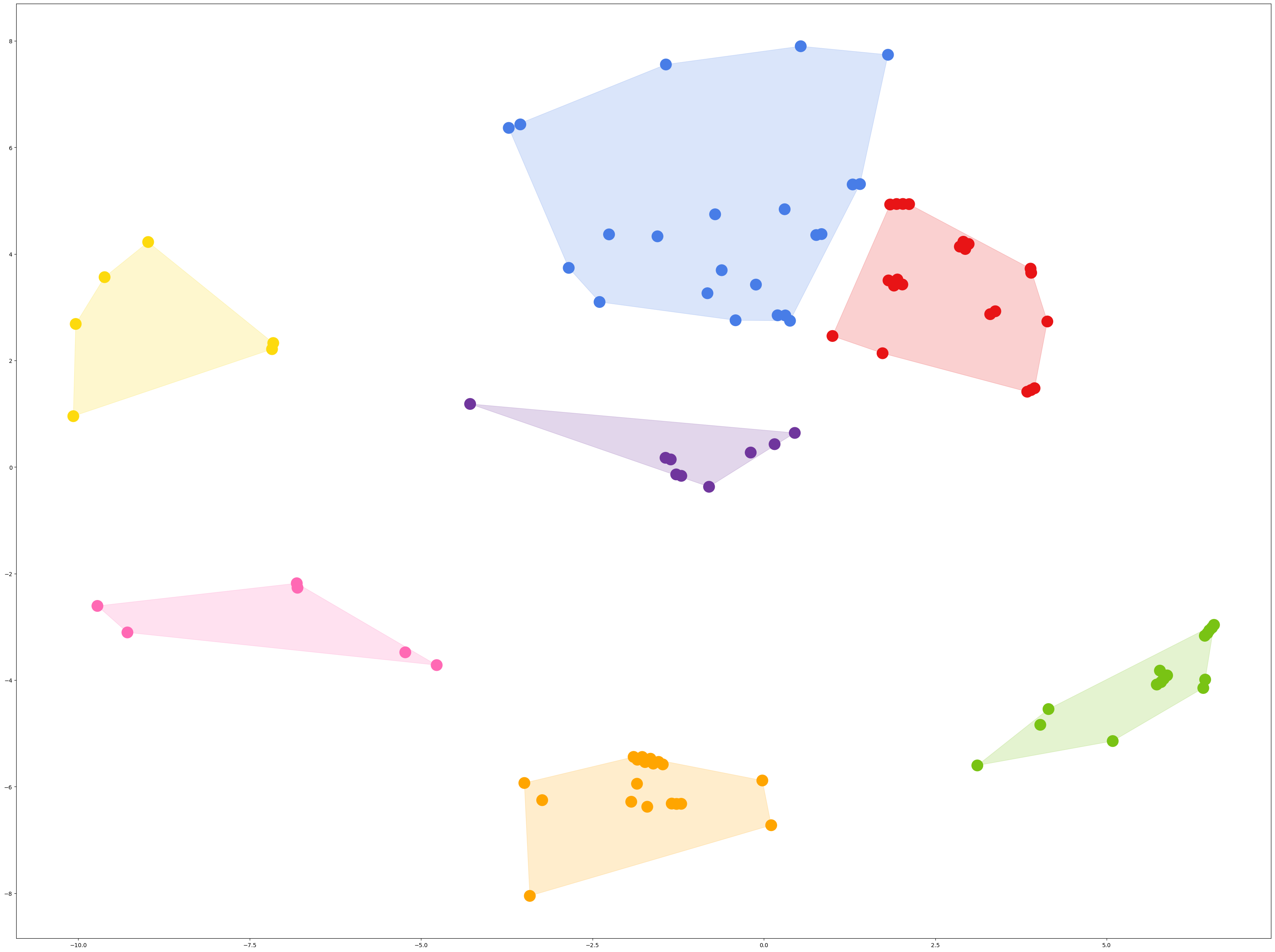}
    \caption{Clusters by MDS using the task distance matrix with the rankings from our survey.}
    \label{fig:cluster-ours}
  \end{subfigure}
  \hfill
    \begin{subfigure}{0.32\linewidth}
    \includegraphics[width=0.95\textwidth,height=2in]{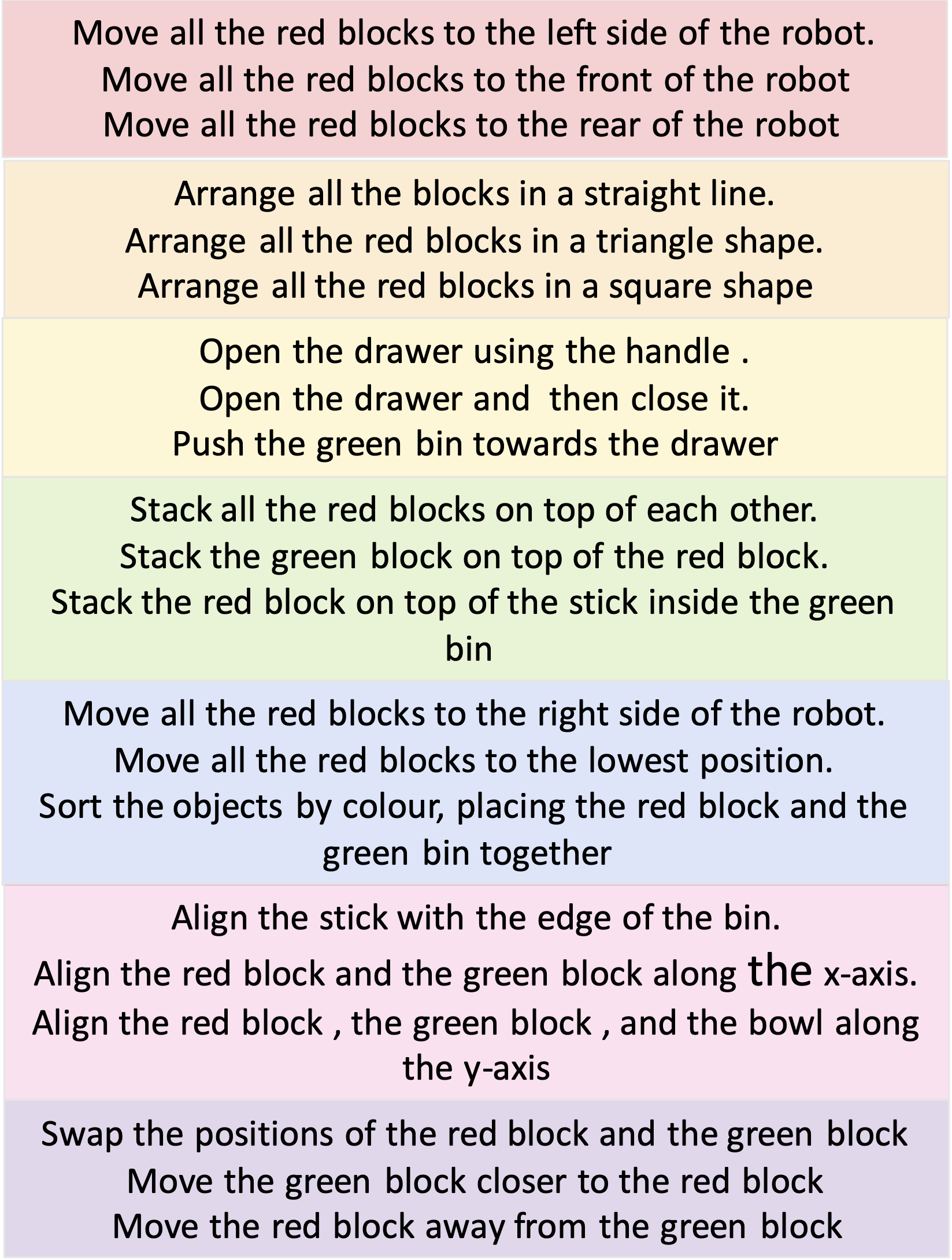}
    \caption{Exemplar tasks from corresponding clusters in (a). Enlarge for details.}
    \label{fig:task-list-clusters}
  \end{subfigure}
  \hfill
  \begin{subfigure}{0.32\linewidth}
    \includegraphics[width=0.95\textwidth,height=2in]{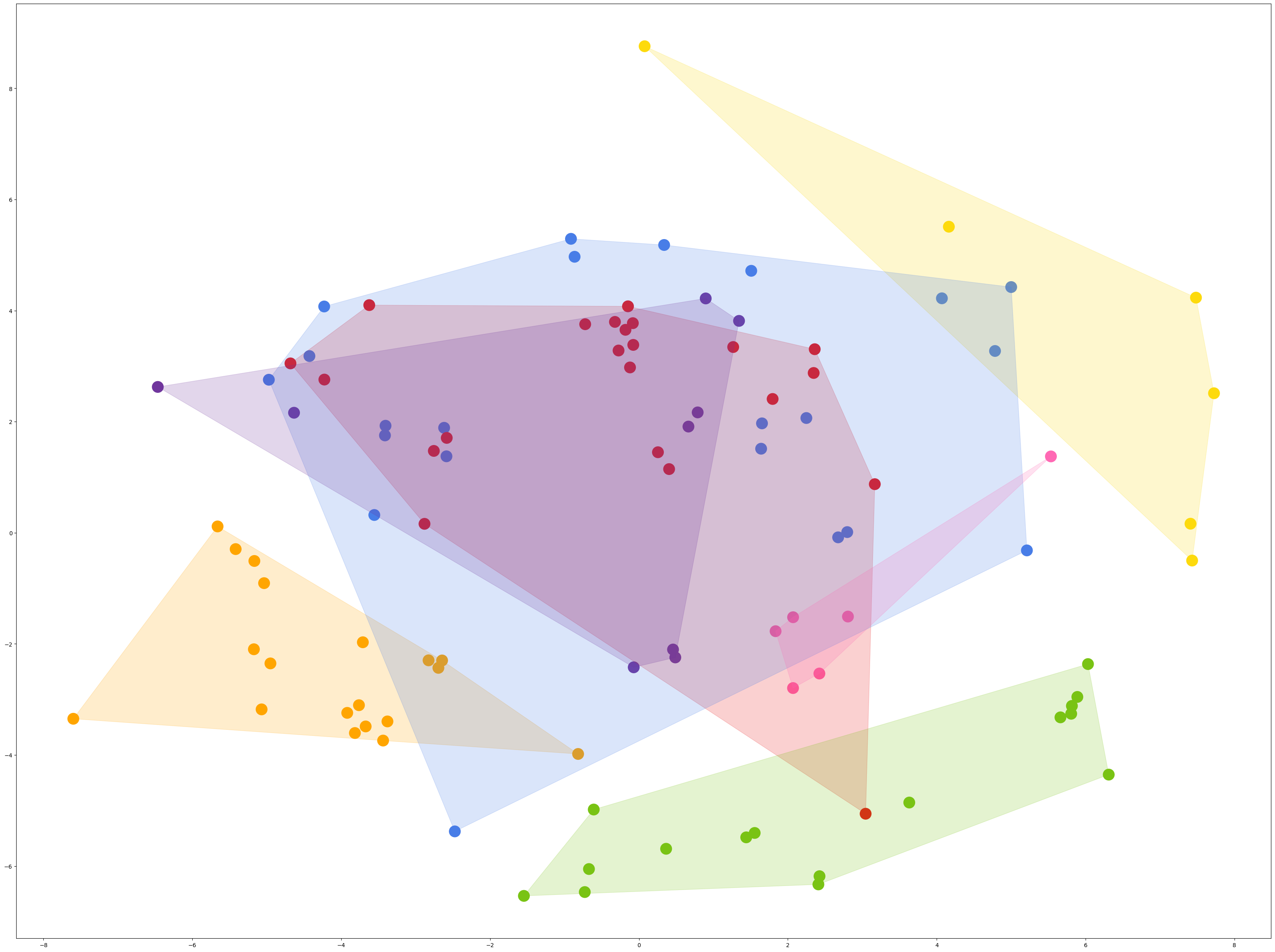}
    \caption{Same clusters on MDS plot of text embeddings of the proposed tasks.}
    \label{fig:plot-text-emb}
  \end{subfigure}
  \vspace{-2mm}
  \caption{Multidimensional Scaling (MDS) is performed on the (human-endorsed) distance matrix to obtain a 2D plot of the tasks, which are further clustered by K-Means (left). 
  From the clusters, a few tasks are chosen and highlighted (middle). 
  MDS is performed again but with the text embeddings of the tasks, however, the clusters from Fig.~\ref{fig:cluster-ours} are now overlapped with each other, evidencing that text-embeddings may not characterize the task space compatible with human understanding.}
  \vspace{-2mm}
  \label{fig:cluster-text-emb}
\end{figure*}

\section{Experiments}
\label{sec:exps}

We evaluate the effectiveness of our pipeline in a tabletop manipulation environment provided by \citep{ha2023scalingup}. 
This evaluation is conducted 
via collecting trajectories in a way that mimics the continuous skill acquisition process
and by distilling a multi-task policy based on the proposed tasks.
We design the experiments to answer the following questions: 

\begin{itemize}[leftmargin=*]
    \vspace{-3mm}
    \item Can the proposed synchronization pipeline generate diverse tasks consistent with human understanding of manipulation task diversity?
    \vspace{-2mm}
    \item Is our pipeline reliable in task generation and success inference, and how essential is the scene graph in ensuring the quality of these tasks?
    \vspace{-2mm}
    \item How well does the distilled policy (synchronized body) perform compared to the elementary data collection policy (factory setting)?
    \vspace{-2mm}
    \item Can the policy perform effectively and adapt efficiently to unseen tasks as it accumulates interaction experience of learned tasks?
\end{itemize}

\vspace{-3mm}
\begin{table}[t]
\centering
\scriptsize
  \caption{Feasibility rate (\%) of the tasks proposed by different task proposal methods.}
  \begin{tabular}{lccccc}
    \toprule
    Baselines  & Scene 0 & Scene 1 & Scene 2 & Scene 3 & Scene 4 \\
    \midrule
    BBSEA & \textbf{100.00} & \textbf{100.00} & \textbf{100.00} & \textbf{100.00} & 70.00 \\
    BBSEA w/o bbox & \textbf{100.00} & 60.00 & \textbf{100.00} & 70.00 & \textbf{86.67} \\
    BBSEA w/o bbox & \textbf{100.00} & 70.00 & \textbf{100.00} & 60.00 & 80.00 \\
    \&positions & & \\
    GPT-4V & 50.00 & 62.50 & 80.00 & 60.00 & 30.00 \\
    GPT-4V-SG & 60.00 & 60.00 & 80.00 & 70.00 & 30.00 \\
    \bottomrule
  \end{tabular}
  \vspace{-2mm}
  \label{tab:task-proposal-feasibility}
\end{table}
\vspace{-3mm}
\begin{table}[t]
\centering
\scriptsize
  \caption{Accuracy (\%) of the tasks completion inference by different success detection methods.}
  \begin{tabular}{lccccc}
    \toprule
    Baselines  & Scene 0 & Scene 1 & Scene 2 & Scene 3 & Scene 4 \\
    \midrule
    BBSEA & \textbf{100.00} & \textbf{100.00} & \textbf{100.00} & \textbf{83.34} & \textbf{100} \\
    BBSEA w/o bbox & {83.34} & \textbf{100.00} & 66.66 & \textbf{83.34} & \textbf{100} \\
    BBSEA w/o bbox & 66.67 & 50.00 & 50.00 & 50.00 & 50.00 \\
    \& positions & & \\
    GPT-4V & 50.00 & 33.33 & 50.00 & 0.00 & 33.33 \\
    GPT-4V-SG & 50.00 & 50.00 & 50.00 & 0.00 & 33.33 \\
    \bottomrule
  \end{tabular}
  \vspace{-2mm}
  \label{tab:task-inference-accuracy}
\end{table}

\subsection{Analysis of Task Proposal Diversity}
\label{sec:diversity-analysis}

We first assess our system's capability to generate a contextually relevant array of tasks that are diverse enough by human standards.

\textbf{Experiment Setting.}
First, we propose a scoring metric to quantify the similarity between two tasks based on human rankings of a set of critical factors. 
The rankings are obtained through a survey from a diverse population of subjects (e.g., college students, bankers, doctors, and engineers).
Please refer to Appendix~\ref{sec:appendix-task-deversity-analysis} for details about the scoring metric. 

We apply this scoring metric on a set of 100 tasks proposed by the task proposer module to obtain a distance matrix containing pairwise distances between tasks.

\textbf{Visualization and Result.}
We perform Multidimensional Scaling (MDS) on the distance matrix and cluster the tasks by K-Means, as shown in Fig.~\ref{fig:cluster-ours}. 
MDS is performed again using the GPT text embedding of the task descriptions and we visualize the previously obtained clusters in Fig.~\ref{fig:plot-text-emb}.
The overlapping clusters highlight that the GPT text embedding may not represent the diversity of tasks well regarding human cognition. 
A snapshot of tasks in each cluster is also shown in Fig.~\ref{fig:task-list-clusters}.
This study shows that the tasks proposed by our pipeline are potentially more diverse than those tasks selected according to the text-embedding-based diversity.
\vspace{2mm}

\vspace{-2mm}
\begin{table}[t]
    \centering
    \scriptsize
        \caption{Success rate (\%) of the data collection policy and the distilled policy across eight tasks.}
    \begin{tabular}{@{}lcc@{}}
      \toprule
         \multicolumn{1}{c}{Tasks} & \multicolumn{1}{c}{Data Collection} & \multicolumn{1}{c}{Distilled} \\
      & \multicolumn{1}{c}{Policy} & \multicolumn{1}{c}{Policy} \\
      \midrule
      Push the bowl towards the rear & 40.00 & 100.00 \\
      Close the drawer & 92.50 & 100.00  \\
      Align the red block with the stick & 32.50 & 50.00 \\
      Move the green block into the open drawer & 17.50 & 75.00 \\
      Push the red block towards the drawer  & 42.50 & 65.00  \\
      Push the red block towards the right & 55.00  & 62.50 \\
      Launch the green block using the catapult& 32.50   & 52.50  \\
      Move the red block to the left of the stick & 35.00  & 45.00 \\\cline{1-2}
      \toprule
      Average & 43.44 & \textbf{68.75}\\ 
      \bottomrule
    \end{tabular}
    \vspace{-4mm}
    \label{tab:distilled-eval}
\end{table}

\vspace{-3mm}
\begin{table}[t]
    \centering
    \scriptsize
        \caption{Success rate (\%) of distilled policies trained with 10, 30, and 60 tasks on four unseen tasks (zero-shot).}
    \begin{tabular}{@{}lccc@{}}
      \toprule
      Tasks  & 10-task & 30-task & 60-task\\
      \midrule
      Move the green block into the  & 0.00 & 5.71 & 34.28 \\
      green bin & & \\
      Gather the red block and the green block & 0.00 & 30.00 & 52.50  \\
      into the green bin & & \\
      Push the block towards rear & 0.00 & 27.50 & 50.00 \\
      Move the turbo airplane toy towards the line & 0.00 & 0.00 & 57.50 \\
      \toprule
      Average & 0.00 & 15.80 & \textbf{48.57}\\ 
      \bottomrule
    \end{tabular}
    \vspace{-2mm}
    \label{tab:zero-shot-capability}
\end{table}

\begin{table*}[h]
  \centering
  \scriptsize
    \caption{Fine-tuning performance in success rate (\%) on various unseen tasks. The policies are pre-trained with varying number of tasks (0, 10, 30 and 60 tasks), and then fine-tuned with different number of demos (20, 50, 80, 100) across four unseen tasks.}
  \begin{tabular}{@{}c|cccc|cccc|cccc|cccc@{}}
    \toprule
    \multicolumn{1}{c}{Pre-trained} & \multicolumn{4}{c}{Gather} & \multicolumn{4}{c}{Bus} & \multicolumn{4}{c}{Bin} & \multicolumn{4}{c}{Drawer} \\
    policies& 20 & 50 & 80 & 100 & 20 & 50 & 80 & 100 & 20 & 50 & 80 & 100 & 20 & 50 & 80 & 100 \\
    \midrule
    0-tasks & 11.25 & 8.75 & 13.75 & 15.00
       &13.75 & 26.25 & 27.50 & 36.25
       &18.75 & 26.25 & 37.50 & 46.25
       &6.25 & 1.25 & 2.50 & 5.00\\
    10-tasks & 10.00 & 18.75 & 17.50 & 21.25
       &16.25 & 38.75 & \textbf{40.00} & 40.00
       &11.25 & 22.50 & 22.50 & 41.25
       &0.00 & 0.00 & 0.00 & 2.50 \\
    30-tasks & 28.75 & 33.75 & 35.00 &41.25 
       &22.50 & 40.00 & \textbf{40.00} &42.50
       &18.75 & 28.75 & \textbf{45.00} & 57.50
       &7.50 & 5.00 & 5.00 & 11.50\\
    60-tasks & \textbf{33.75} & \textbf{36.25} & \textbf{42.50} & \textbf{42.50}
       &\textbf{30.00} & \textbf{43.75} & \textbf{40.00} & \textbf{57.50}
       &\textbf{21.25} & \textbf{30.00} & 38.75 & \textbf{58.75}
       &\textbf{8.75} & \textbf{7.50} & \textbf{7.50} & \textbf{15.00}\\
    \bottomrule
  \end{tabular}
  \vspace{-2mm}
  \label{tab:finetuning-eval}
\end{table*}

\vspace{2mm}
\subsection{Analysis of Task Proposal Feasibility}
\label{sec:feasibility-analysis}

We then conduct an evaluation of the feasibility of proposed tasks from the task proposer. 
We consider the baselines of different task proposal methods: (1) \textbf{BBSEA} queries GPT-4 with full information of the scene graph. 
(2) \textbf{BBSEA w/o bbox} queries GPT-4 without bounding boxes of the objects in the scene graph. 
(3) \textbf{BBSEA w/o bbox\&positions} queries GPT-4 without bounding boxes and positions of the objects in the scene graph.
(4) \textbf{GPT-4V} queries GPT-4V with front view image, without any information of the scene graph.
(5) \textbf{GPT-4V-SG} queries GPT-4V with front view image, and is prompted to formulate its own scene graph for task proposal.
See Appendix~\ref{sec:prompts for large language model} for the full list of prompts.
The evaluation is conducted across five randomly chosen scenes (the environments are visualized in Appendix~\ref{sec:appendix-training-details}).
For each baseline, we query GPT-4/4V in multiple trials and assess the feasibility based on the proportion of the tasks that can be used to collect successful demos for policy training.

Results in Tab.~\ref{tab:task-proposal-feasibility} reveal that tasks generated by our pipeline demonstrate significantly higher feasibility compared to those proposed by GPT-4V/4V-SG, which matches the case shown in Fig.~\ref{fig:task-proposal-ours-vs-gpt4v}. 
It shows the imprecise and ungrounded understanding of GPT-4V and the need for our robust sensing module. 
Also, the ablations of scene graph (w/o bbox and w/o bbox\&positions) are worse than the original pipeline, which shows that the usage of scene graph is essential in ensuring the quality of the proposed tasks.

\subsection{Analysis of Success Inference Accuracy}
\label{sec:inference-analysis}

We further investigate the accuracy of our task completion inference module. 
The baseline methods and evaluation scenes from the previous section \ref{sec:feasibility-analysis} are retained.
This analysis focuses on the ability of these methods to accurately infer whether a proposed task has been successfully completed, which is crucial for effective demonstration collection and policy training in our pipeline. 

The results in Tab.~\ref{tab:task-inference-accuracy} reveal that our BBSEA framework, which leverages complete scene graph information, is more accurate than other approaches when inferring task success.
The consistent high scores observed for the BBSEA framework across various scenes demonstrates the importance of detailed spatial data, including bounding boxes and object positions. 
In contrast, GPT-4V and GPT-4V-SG which rely solely on visual inputs, show lower accuracy. 
The findings underscore the vital role of comprehensive scene graph information for precise success inference.

\subsection{Evaluation of Distilled Policies}
\label{sec:evaluation-distilled-policy}

We evaluate the performance of the policies learned with BBSEA by comparing it against the policy we used to collect data.
This assessment aims to demonstrate the necessity and effectiveness of the policy distillation.

\textbf{Experiment Setting.} 
We train a multi-task policy on eight tasks with 200 demonstrations per-task. 
The implementation details including environment and training settings are shown in Appendix~\ref{sec:appendix-training-details}.
The performance of distilled policy is then evaluated on the eight tasks, and compared with the original data collection policy.

\textbf{Main Results.} The results in Tab.~\ref{tab:distilled-eval} reveal an improvement by 25\% in average success rate of the distilled policy over the data collection policy, affirming the effectiveness of our methodology in skill acquisition and policy evolution.

\subsection{Zero-shot Capability Assessment}
\label{sec:evaluation-zero-shot}
We then assess the zero-shot capability of our pipeline by evaluating tasks not seen during training. 
Following the training procedure in Sec.~\ref{sec:evaluation-distilled-policy}, we evaluate the policies trained on 10, 30, and 60 proposed tasks on four unseen tasks respectively. 
The task details are shown in Appendix~\ref{sec:appendix-task-details}.
We also visualize the MDS diversity of these tasks in Fig.~\ref{fig:area-span-comparison-all}, suggesting that a larger task set contributes to a broader range of task variations.
The results of our assessment are presented in Tab.~\ref{tab:zero-shot-capability}, which indicate that policies trained on a larger variety of tasks can exhibit better zero-shot generalization. 
This suggests that the breadth of training directly influences the system's ability to adeptly handle tasks it has not encountered before, highlighting the significance of scaling diverse skills using our automatic pipeline.

\subsection{Adaptation Performance on Novel Tasks}
\label{sec:fine-tuning-new-tasks}

We further evaluate the adaptation performance of our derived policies across four unseen tasks: ``gather'', ``bin'', ``bus'' and ``drawer''.
We choose the distilled policies (trained on 10, 30, and 60 tasks) from Sec.~\ref{sec:evaluation-zero-shot} as pre-trained policies. 
We then fine-tune the policies across the four unseen tasks with varying demonstrations per task.
The outcome in Tab.~\ref{tab:finetuning-eval} demonstrates that policies trained with a larger task variety exhibit better performance after fine-tuning. 
Also, distilled policies trained on 60 tasks show the highest success rate compared to others. 
Despite small variations induced by the training dynamics, our experiments confirm the importance of acquiring physical skills across diverse tasks. 
This emphasizes the necessity of autonomously proposing and learning interaction policies as in the proposed BBSEA pipeline.

%% file: sec/5_conclusion.tex
\section{Conclusion}
\label{sec:conclusion}

We present a brain-body synchronization pipeline to enable embodied agents to autonomously acquire various interaction skills without human intervention. 
Leveraging large foundation models, our pipeline proposes tasks for learning and establishes success metrics for providing learning feedback. 
Our experiments in the widely adopted tabletop settings demonstrate that with a structured representation of scene information and a small set of in-context examples, a collection of large foundation models (representing the {\it brain}) can effectively suggest scene-compatible and physically feasible tasks for an agent ({\it body}).

This approach facilitates autonomous skill acquisition, 
enabling the training of language-conditioned policies that exhibit promising zero-shot and few-shot generalization capabilities on novel tasks and environments. 

While our experiments provide valuable insights, 
they are primarily confined to tabletop environments and utilize a limited set of action primitives. 
These are the current limitations of our approach for dealing with complicated real-world scenarios. 
In future work, we aim to extend our pipeline beyond the boundaries, exploring more intricate and realistic environments and integrating a wider array of action primitives. 
This extension is essential for fully realizing the potential of our methods and for advancing the application of autonomous skill acquisition in more diverse and complex scenarios.


\section{Acknowledgment}

This work is supported by the HKU-100 Award, the Microsoft Accelerate Foundation Models Research Program, and in part by the JC STEM Lab of Robotics for Soft Materials funded by The Hong Kong Jockey Club Charities Trust.

%% file: sec/X_suppl.tex
\newpage
\appendix
\section*{Appendix}

\section{Implementation Details}
\label{sec:appendix-implementation-details}
In this section, we describe the implementation details of some components of our pipeline. Please visit \href{https://bbsea-embodied-ai.github.io}{https://bbsea-embodied-ai.github.io} for more visualizations.

\subsection{Scene Comprehension}
\label{sec::appendix-implementation-details-scene-comprehension}
\paragraph{Object Detection and Segmentation}
In our pipeline, the process of scene comprehension starts with detecting the objects in the images obtained via RGBD cameras. 
The object detector \citep{Jocher_YOLOv5_by_Ultralytics_2020} outputs 2D bounding boxes and names of the detected objects. 
Then SAM \citep{SAM} is utilized to refine the spatial occupancy of the detected objects and to generate segmentation masks, which enables the extraction of corresponding point clouds for each object given the camera parameters and depth. 
This process is equally feasible in the real world. 
Our experiment utilized a simulated environment, and for convenience, we directly obtained semantic segmentation information in the images captured by the camera in the simulator.

\paragraph{Object State Detection}
\label{sec::appendix-implementation-details-object-state-detection}
The state of objects is crucial for task proposal and for inferring whether the task has been accomplished. 
For objects with different intrinsic states, we pre-define lists of semantically meaningful states depending on their categories. 
The full lists of the concerned states for different objects are shown in Listing \ref{listing:states-of-objects}.
\begin{lstlisting}[language=Python, caption=Lists of the concerned states of different objects., label=listing:states-of-objects, frame=none, basicstyle=\small\ttfamily, commentstyle=\color{citecolor}\small\ttfamily,columns=fullflexible, breaklines=true, postbreak=\mbox{\textcolor{red}{$\hookrightarrow$}\space}, escapeinside={(*}{*)}]
STATE_DICT = {
  'drawer': ['open', 'closed'],
  'catapult': ['triggered', 'not triggered']
}
\end{lstlisting}

\subsection{Primitive Actions}
\label{sec::appendix-implementation-details-primitive-actions}
The primitive actions consist of a waypoint sampler and a motion planner. 
The waypoint sampler samples potential waypoints that indicate gripper poses and open/closed states. 
The motion planner figures out configurations of the robot arm that could achieve the sampled waypoints.

Currently, we compose a feasible set of actions for the robot arm using seven primitive actions, including ``Pick(obj\_name)'', ``PlaceOn(obj\_name)'', ``PlaceAt(place\_pos'', ``Push(obj\_name, direction, distance)'', ``PrismaticJointOpen(obj\_name)'', ``PrismaticJointClose(obj\_name)'' and ``Press(obj\_name)''. 
These primitive actions are sufficient to accomplish a wide range of tasks. 
Further expanding the set of primitive actions will be considered for future work.

Listing \ref{listing:explanation-of-the-primitive-actions} illustrates the functionality and parameters of the primitive actions, with specific implementation details referenced in the code of our pipeline, which we are committed to releasing.

\begin{lstlisting}[caption=Explanation of the primitive actions., label=listing:explanation-of-the-primitive-actions, frame=none, basicstyle=\small\ttfamily, commentstyle=\color{citecolor}\small\ttfamily, breaklines=true, escapeinside={(*}{*)}]
Pick(obj_name): Approach the object, close the gripper to grasp it and lift it up. (Parameters: obj_name -- the name of the object which would be picked)
PlaceOn(obj_name): Move the gripper on top of the object with another object in the gripper and then open the gripper. (Parameters: obj_name -- the name of the object which an object in the gripper would be placed on)
PlaceAt(place_pos): Move the gripper to the target position with an object in the gripper and then open the gripper. (Parameters: place_pos -- the target position which an object in the gripper would be moved to)
Push(obj_name, direction, distance): Close the gripper and then push the object in the specified direction by a specified distance. (Parameters: obj_name -- the name of the object which would be pushed; direction -- the direction which the object would be moved in, it is direction vector [x, y]; distance -- the distance which the object would be moved by, the distance is in the unit of meter)
PrismaticJointOpen(obj_name): Open the object with a prismatic joint. (Parameters: obj_name -- the name of the handle of the object with a prismatic joint)
PrismaticJointClose(obj_name): Close the object with a prismatic joint. (Parameters: obj_name -- the name of the handle of the object with a prismatic joint)
Press(obj_name): Close the gripper and then press the object. (Parameters: obj_name -- the name of the object which should be pressed)
\end{lstlisting}

\subsection{Language Conditioned Multi-Task Policy}
\label{sec::appendix-implementation-details-language-conditioned-multi-task-policy}
We utilize the official code of visuomotor policy in \citep{ha2023scalingup} which is available at \href{https://github.com/real-stanford/scalingup}{github.com/real-stanford/scalingup} as the implementation of the language conditioned multi-task policy. Further details can be found there.

\section{Task Diversity Analysis}
\label{sec:appendix-task-deversity-analysis}
A questionnaire is designed by identifying a set of 5 factors which would impact how we quantify the difference between two tasks. These factors are identified as follows:
\begin{enumerate}
    \item The main action involved in the tasks.
    \item The object on which the action is implemented in the tasks
    \item The shape of the location mentioned in the tasks
    \item The color of the object of focus in the tasks
    \item The color of the target location in the tasks
\end{enumerate}
The human perception of the importance of factors is subjective but a survey can provide a starting point regarding the ranking of these factors. To identify the rankings, we asked a group of people from various backgrounds a set of questions where each question consisted of two pairs of tasks. For example, Set A consists of tasks which differ in the main actions. Set B consists of tasks which differ by only one factor - the object used. Similarly Set C has tasks with different target locations, Set D has tasks with different colors of the objects and Set E has tasks where the color of the target is different.

\vspace{0.05in}
\textbf{Questionnaire Sample:} 
\begin{enumerate}
\item Given the two sets of tasks, which set do you think has a larger distance/difference between them ?\\
\textbf{Set A}\\
Push the green bin close to the drawer\\
Pick up and place the green bin close to the drawer.\\
\textbf{Set B}\\
Push the green bin close to the drawer \\
Push the green bowl close to the drawer
\item Given the two sets of tasks, which set do you think has a larger distance/difference between them ?\\
\textbf{Set A}\\
Push all the red blocks in a triangle shape.\\
Pick up and place all the red blocks in a triangle shape.\\
\textbf{Set C}\\
Pick up and place all the red blocks in a triangle shape. \\
Pick up and place all the red blocks in a circular shape.
\item Given the two sets of tasks, which set do you think has a larger distance/difference between them ?\\
\textbf{Set A}\\
Pick up and place the red block next to the green bowl. \\
Push the red block next to the green bowl.\\
\textbf{Set D}\\
Push the red block next to the green bowl. \\
Push the blue block next to the green bowl.
\item Given the two sets of tasks, which set do you think has a larger distance/difference between them ?\\
\textbf{Set A}\\
Pick up and place all the green blocks in the green square. \\
Push all the green blocks in the green square.\\
\textbf{Set E}\\
Push all the green blocks in the green square. \\
Push all the green blocks in the yellow square.
\item Given the two sets of tasks, which set do you think has a larger distance/difference between them ?\\
\textbf{Set B}\\
Move the block into the drawer \\
Move the ball into the drawer\\
\textbf{Set C}\\
Move the block into the drawer \\
Move the block into the bin
\item Given the two sets of tasks, which set do you think has a larger distance/difference between them ?\\
\textbf{Set B}\\
Pick and place the red airplane toy into the green bin. \\
Pick and place the red block into the green bin.\\
\textbf{Set D}\\
Pick and place the red block into the green bin. \\
Pick and place the green block into the green bin.
\item Given the two sets of tasks, which set do you think has a larger distance/difference between them ?\\
\textbf{Set B}\\
Push the block towards the green drawer. \\
Push the bowl towards the green drawer. \\
\textbf{Set E}\\
Push the block towards the green drawer. \\
Push the block towards the blue drawer.
\item Given the two sets of tasks, which set do you think has a larger distance/difference between them ?\\
\textbf{Set C}\\
Pick and place all the red blocks in a bowl.\\
Pick and place all the red blocks in a square shape.\\
\textbf{Set D}\\
Pick and place all the red blocks in a bowl. \\
Pick and place all the green blocks in a bowl.
\item Given the two sets of tasks, which set do you think has a larger distance/difference between them ?\\
\textbf{Set C}\\
Move all the green bowls to the cupboard \\
Move all the green bowls to the green square.\\
\textbf{Set E}\\
Move all the green bowls to the green square. \\
Move all the green bowls to the blue square.
\item Given the two sets of tasks, which set do you think has a larger distance/difference between them ?\\
\textbf{Set D}\\
Swap the positions of the red block and the green block \\
Swap the positions of the blue block and the green block.\\
\textbf{Set E}\\
Swap the positions of the red block and the green block \\
Swap the positions of the red block and the blue block.
\end{enumerate}

\vspace{0.1in}

Each set corresponds to a specific factor and all the factors are compared against each other. We count the number of responses in favour of each set (Tab.~\ref{tab:survey-response}) and rank them.

\begin{table}[h]
    \centering
        \captionof{table}{Responses collected and the corresponding votes for each factor which makes the tasks more distant.}
    \begin{tabular}{lc}
      \toprule
     Factor & Total Count \\
      \midrule
      Set A \textbf{ - Main Action} & 166 \\
      Set B \textbf{ - Shape of Object} & 112 \\
      Set C \textbf{ - Shape of Location} & 130\\
      Set D \textbf{ - Object color} & 53 \\
      Set E \textbf{ - Location color} & 39 \\\cline{1-2}
      \toprule
      Total  &  \textbf{500}\\ 
      \bottomrule
    \end{tabular}
    \vspace{-2mm}
    \label{tab:survey-response}
\end{table}

\textbf{Scoring System for Distance Matrix:}
The distance between tasks numbered $i$ and $j$ is defined as follows: 
\vspace{-2mm}
\begin{equation*}
\text{distance}[i, j] = \sum_{k=1}^{5} \text{score}_k(i,j)
\vspace{-2mm}
\end{equation*}
where:
\begin{itemize}
    \vspace{-3mm}
    \item $\textnormal{score}_1$ depends on the disparity in the main action of the tasks. A score of 5 is assigned if the actions are different, and 0 if they are the same.
    \vspace{-2mm}
    \item $\textnormal{score}_2$ depends on the disparity in the shape of the target locations in the tasks. A score of 4 is assigned if the shapes are different, and 0 otherwise.
    \vspace{-2mm}
    \item Similarly, $\textnormal{score}_3$ depends on the disparity in the shape of the objects in the tasks. A score of 3 is assigned if the shapes are different.
    \vspace{-2mm}
    \item $\textnormal{score}_4$ depends on the disparity in the color of the objects in the tasks. A score of 2 is assigned if not the same.
    \vspace{-2mm}
    \item $\textnormal{score}_5$: depends on the disparity in the color of the target in the tasks. A score of 1 is assigned if they are different.
\end{itemize}

A higher score between tasks numbered $i$ and $j$ implies that the tasks are more different from each other as compared to the case when the score between two tasks is lower. We refer to this score as the distance between the tasks. 

\begin{figure*}[ht]
  \centering
  \includegraphics[width=0.82\linewidth]{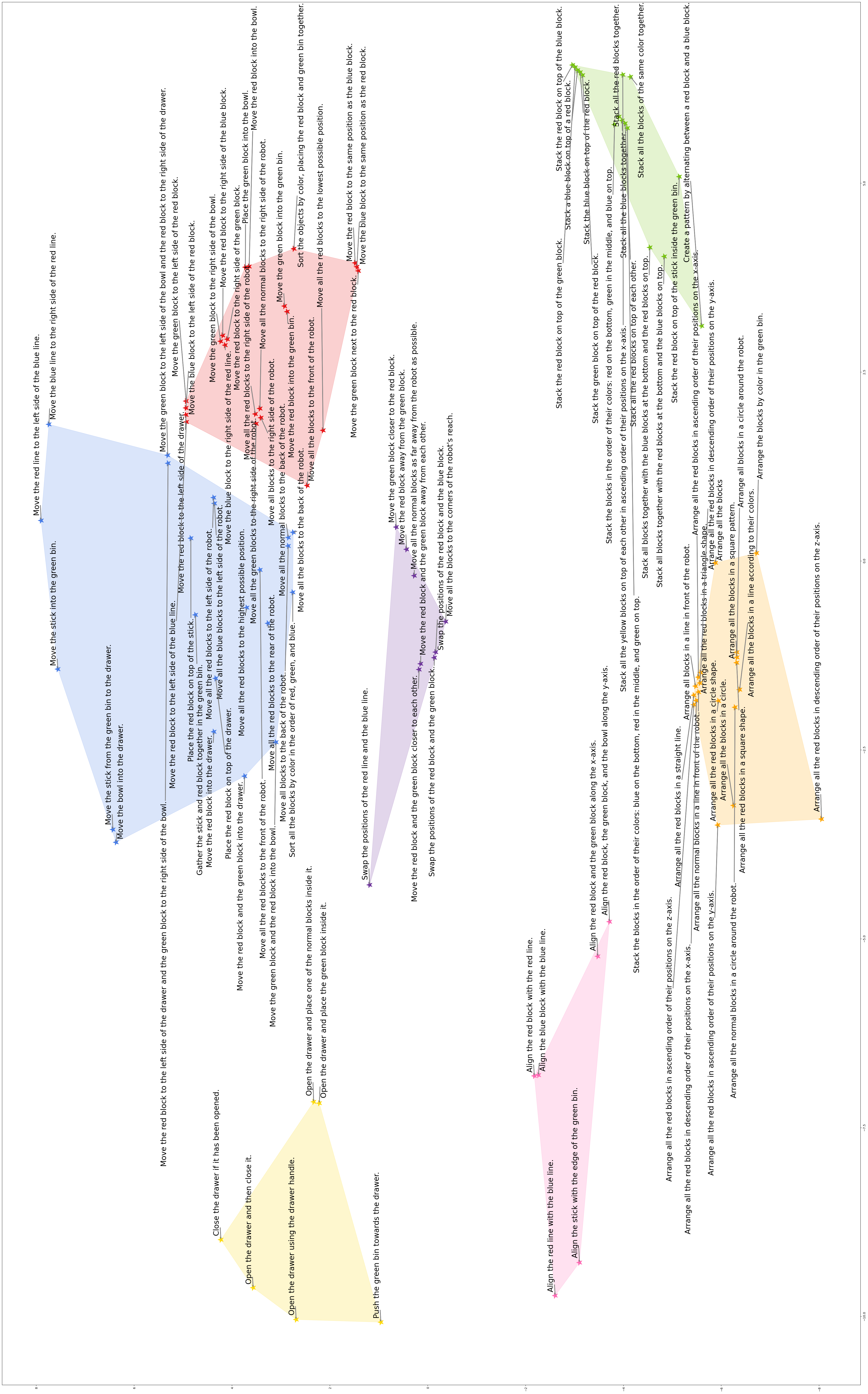}
    \vspace{-6mm}
  \caption{An overview of the MDS plot obtained with all 100 tasks from the task proposal module which are clustered using K-Means Clustering. Please enlarge to see details of the proposed tasks.}
  \label{fig:alltask-list-MDS}
  \vspace{-22.81828pt}
\end{figure*}

\textbf{Visualization:} To obtain the multidimensional scaling (MDS) plot , we use python's sklearn library. After obtaining pairwise distances for a set of tasks, the symmetric matrix obtained then undergoes MDS so that we can obtain a two dimensional plot to visualize the distance between the tasks and proceed to cluster them. The clusters obtained along with the tasks are shown in Fig.~\ref{fig:alltask-list-MDS}.

\begin{figure*}[t]
  \centering
  \includegraphics[width=0.65\linewidth]{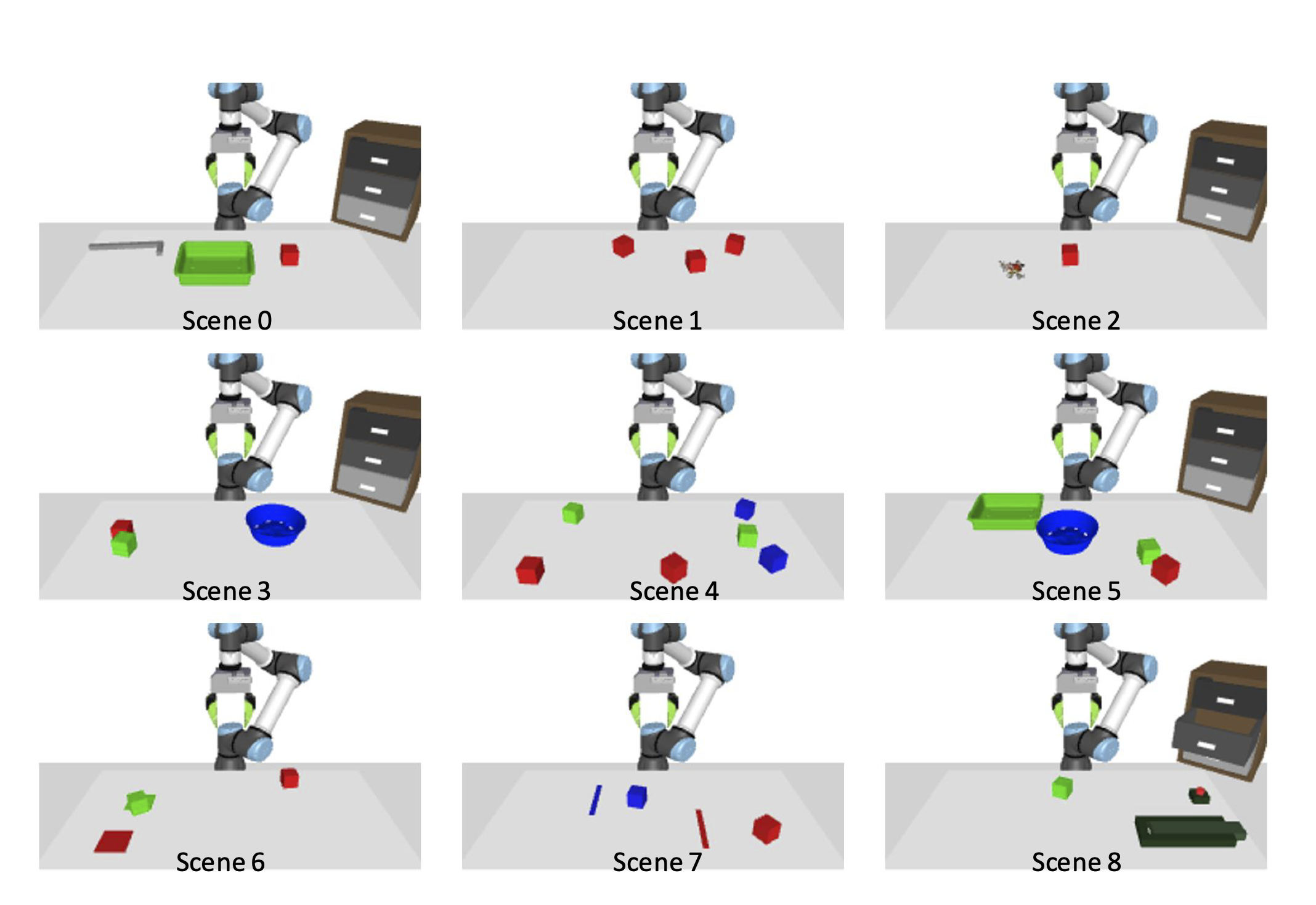}
  \caption{Visualization of the environments in which brain-body synchronization is performed on.}
  \vspace{-1mm}
  \label{fig:env-vis}
\end{figure*}

\section{Environment and Training Details}
\label{sec:appendix-training-details}
We train multi-task visuomotor policies proposed by \citep{ha2023scalingup} on the data collected by the brain-body synchronization in nine environments. A visualization of the environments is provided in Fig.~\ref{fig:env-vis}. We employ identical network architecture and hyperparameters as the official implementation during training time. The policies are trained with varying task quantities (10, 30, 60), and 200 demonstrations are collected for each task. The corresponding tasks are illustrated in Appendix~\ref{sec:appendix-task-details}. During finetuning in Sec.~\ref{sec:fine-tuning-new-tasks}, we set the epoch to one and the batch size to 256 for rapid adaptation on a small amount of data. The tasks which policies are trained on and generalize to are visualized on our website \href{https://bbsea-embodied-ai.github.io}{https://bbsea-embodied-ai.github.io}.

\section{Task Details}
\label{sec:appendix-task-details}
The policies are trained on 10, 30, and 60 tasks. Below are these tasks which are all proposed by GPT-4 after understanding the scene. The visualization of the trajectories collected on these tasks is available on our website \href{https://bbsea-embodied-ai.github.io}{https://bbsea-embodied-ai.github.io}. For comparison purposes, tasks from previous literature \citep{ha2023scalingup} is also mentioned. In addition to the task details, an analysis of the diversity present in the 4 sets of tasks used for policy training is also presented in this section.\\

\textbf{The 10 tasks:} 
\begin{itemize}
  \item move the green block into the bowl
  \item move the red block away from the bowl
  \item move the green block to the left side of the red block
  \item push the bowl towards the rear
  \item push the red block towards the right
  \item push the green block towards the left
  \item push the red block towards the front
  \item place the red block next to the stick
  \item gather the stick and the red block together near the green bin
  \item align the red block and the stick along the x-axis.
\end{itemize}

\vspace{0.05in}

\textbf{The 30 tasks:} 
\begin{itemize}
  \item move the green block into the bowl
  \item move the red block away from the bowl
  \item move the green block to the left side of the red block
  \item push the bowl towards the rear
  \item push the red block towards the right
  \item push the green block towards the left
  \item push the red block towards the front
  \item place the red block next to the stick
  \item gather the stick and the red block together near the green bin
  \item align the red block and the stick along the x-axis
  \item move the stick into the green bin
  \item gather the stick and the red block into the green bin
  \item open the drawer
  \item stack the red block on top of the stick
  \item move the red block closer to the drawer
  \item push the red block towards the drawer
  \item align the red block with the stick
  \item move the stick to the position of the red block
  \item move the red block away from the drawer
  \item move the stick and the red block to the same y-coordinate
  \item stack all the red blocks on top of each other
  \item arrange all the red blocks in a straight line
  \item gather all the red blocks together
  \item arrange all the red blocks in a triangle shape
  \item move a red block to the left side of the robot
  \item move a red block to the rear of the robot
  \item arrange the red blocks in a diagonal line from the front left to the back right
  \item move all the red blocks to the corner
  \item move the red block to the left of the stick
  \item open the drawer and place the red block inside it
\end{itemize}

\vspace{0.05in}

\textbf{The 60 tasks:} 
\begin{itemize}
  \item move the green block into the bowl
  \item move the red block away from the bowl
  \item move the green block to the left side of the red block
  \item push the bowl towards the rear
  \item push the red block towards the right
  \item push the green block towards the left
  \item push the red block towards the front
  \item place the red block next to the stick
  \item gather the stick and the red block together near the green bin
  \item align the red block and the stick along the x-axis
  \item move the stick into the green bin
  \item gather the stick and the red block into the green bin
  \item open the drawer
  \item stack the red block on top of the stick
  \item move the red block closer to the drawer
  \item push the red block towards the drawer
  \item align the red block with the stick
  \item move the stick to the position of the red block
  \item move the red block away from the drawer
  \item move the stick and the red block to the same y-coordinate
  \item stack all the red blocks on top of each other
  \item arrange all the red blocks in a straight line
  \item gather all the red blocks together
  \item arrange all the red blocks in a triangle shape
  \item move a red block to the left side of the robot
  \item move a red block to the rear of the robot
  \item arrange the red blocks in a diagonal line from the front left to the back right
  \item move all the red blocks to the corner
  \item move the red block to the left of the stick
  \item open the drawer and place the red block inside it
  \item move the blocks on the corresponding colored stickers
  \item move the red block to a different location
  \item stack all the green blocks together
  \item move all the red blocks to the left side of the robot
  \item arrange all the blue blocks in a line
  \item stack all blocks of the same color together
  \item move all the red blocks to the left of the green blocks
  \item move all the green blocks on top of the red blocks
  \item stack three blocks of different colors
  \item stack two blocks of different colors
  \item move the blue block and the red block on opposite ends of the blue line
  \item place the blue block and the red block side by side on the blue line
  \item place the blocks on both sides of the blue line
  \item move the blue block and the red block to the middle of the blue line and the red line respectively
  \item move all the blue blocks to the front and all the red blocks to the back
  \item move the turbo airplane toy to the right side of the red block
  \item gather the turbo airplane toy and the red block near the drawer
  \item move the turbo airplane toy to a position that is closer to the drawer than the red block
  \item pick up the red block
  \item swap the positions of the red block and the turbo airplane toy
  \item sort the objects by moving the red block and the stick into the drawer and the green bin respectively
  \item move the green block into the open drawer
  \item press the catapult button to trigger the catapult
  \item close the drawer
  \item launch the green block using the catapult
  \item align the red block and the turbo airplane toy along the y-axis
  \item move the turbo airplane toy to the front and the red block to the back
  \item arrange the red block and the turbo airplane toy side by side
  \item place the green block on the catapult
  \item move the green block next to the catapult
\end{itemize}

\textbf{Tasks from previous work (Scaling Up and Distilling Down):} 
\begin{itemize}
  \item Move the package into the mailbox and raise the flag
  \item Move the toy from the right bin to the left bin
  \item Move the vitamin bottle into the top drawer
  \item Move the vitamin bottle into the middle drawer
  \item Move the vitamin bottle into the bottom drawer
  \item Move the pencil case into the top drawer
  \item Move the pencil case into the middle drawer
  \item Move the pencil case into the bottom drawer
  \item Move the crayon box into the top drawer
  \item Move the crayon box into the middle drawer
    \item Move the crayon box into the bottom drawer
  \item Move the horse toy into the top drawer
  \item Move the horse toy into the middle drawer
  \item Move the horse toy into the bottom drawer
  \item Move the block into the catapult arm, press the button to shoot the block into the closest bin
  \item Move the block into the catapult arm, press the button to shoot the block into the middle bin
  \item Move the block into the catapult arm, press the button to shoot the block into the furthest bin
  \item Balance the bus toy on the block
\end{itemize}

\vspace{0.05in}

\subsection{Task Diversity for Policy Learning}
\label{sec:appendix-task-diversity-policy-learning}
\begin{figure*}[hb]
  \centering
  \includegraphics[width=0.85\linewidth]{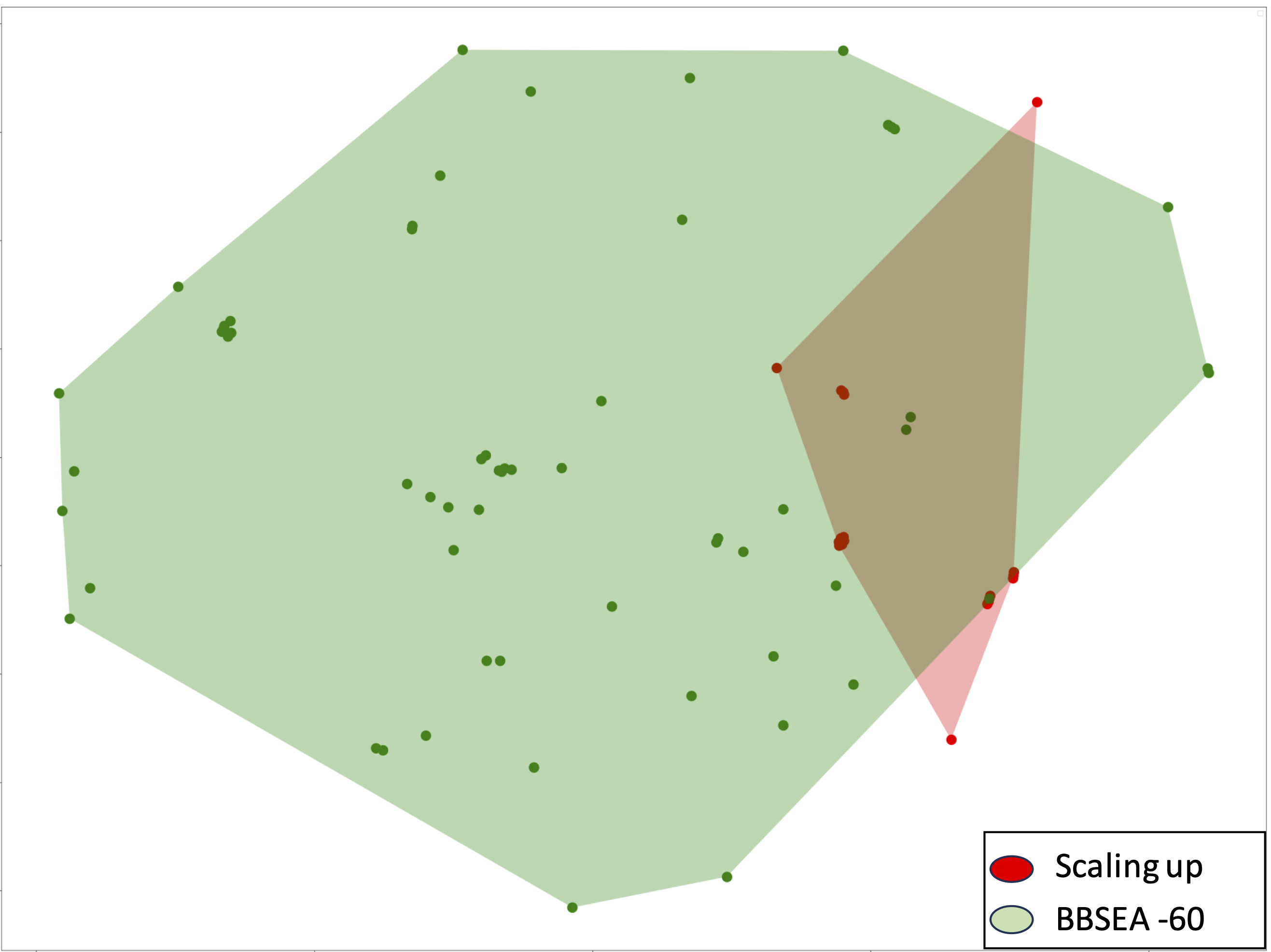}
  \caption{An overview of the MDS plot, which showcases the area covered by the tasks used in BBSEA for policy training (highlighted in green) compared to the tasks previously utilized for policy training in the literature~\citep{ha2023scalingup} (marked in red). The broader expanse of the green cluster indicates that our tasks are more spread out and exhibit greater diversity, according to the our scoring system.}
  \vspace{-1mm}
  \label{fig:area-span-comparison}
\end{figure*}

\begin{figure*}[ht]
  \centering
  \includegraphics[width=0.85\linewidth]{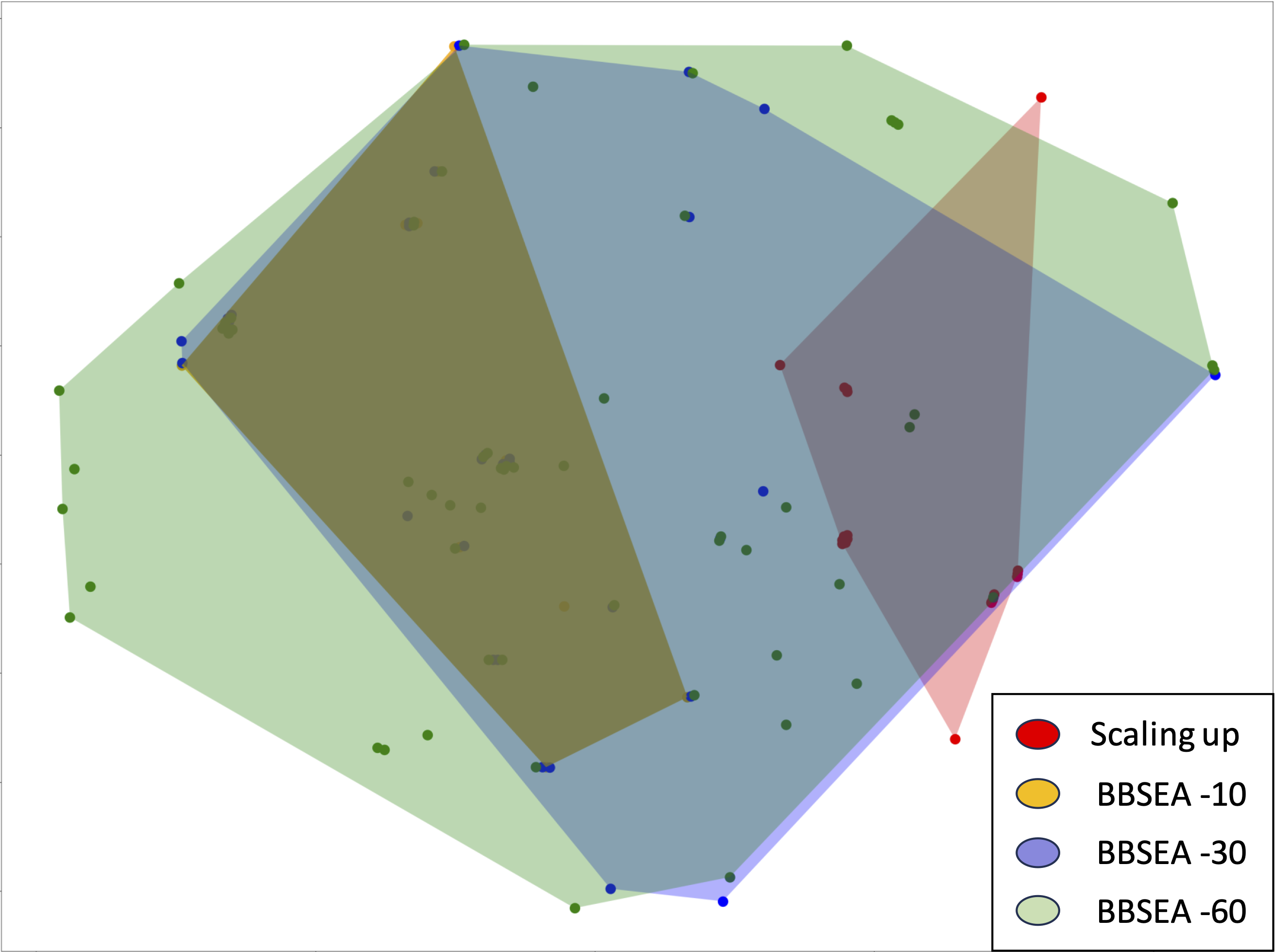}
  \caption{An overview of the MDS plot that highlights the area covered by the 60 tasks utilized in BBSEA for policy training (illustrated in green) in comparison to the tasks previously used for policy training in existing literature~\cite{ha2023scalingup} (depicted in red). Additionally, we showcase the areas spanned by 10 tasks (in orange) and 30 tasks (in blue) used in BBSEA for policy training. The expanded area covered by the BBSEA clusters demonstrates that our tasks are more widely dispersed and exhibit a higher level of diversity, as per our developed scoring system. Furthermore, it is observed that the area covered by the 60 tasks significantly surpasses the areas covered by the 30 and 10 tasks, indicating a greater degree of diversity.}
  \vspace{-1mm}
  \label{fig:area-span-comparison-all}
\end{figure*}

Evaluation of policies is done on a set of 10, 30 and 60 proposed tasks as mentioned in Appendix~\ref{sec:appendix-training-details} and Appendix~\ref{sec:appendix-task-details}. 
To compare the diversity in the tasks on which policies are trained on, we collect our 60 tasks along with the tasks obtained from previous work \citep{ha2023scalingup} and perform similar analysis as mentioned in Sec.~\ref{sec:diversity-analysis} where our scoring system ,as shown in Appendix~\ref{sec:appendix-task-deversity-analysis} ,is used to obtain pairwise distances.
We then perform MDS to obtain a two dimensional plot with the tasks plotted as points. 
The total area spanned by our tasks on the MDS plot covers a larger area as compared to the tasks which were used in previous work.\citep{ha2023scalingup} as shown in Fig.~\ref{fig:area-span-comparison}. 
The MDS plot obtained is a result of the distance matrix from a pairwise distance comparison between all the tasks. A larger distance implies that the tasks are more different from each other in comparison to the tasks which have a smaller distance between them. 
Since the area spanned by the tasks mentioned in BBSEA covers a larger area, we can conclude that the distances between the tasks are larger and hence they are more diverse. 
Furthermore, we performed the same experiment but this time, the 10 tasks used for policy learning in BBSEA and the 30 tasks used for policy learning were also included, and the areas covered by the sets of tasks are highlighted as shown in Fig.~\ref{fig:area-span-comparison-all} which shows that the 60 tasks used are more diverse than the cases where 10 and 30 tasks are used in the BBSEA framework.

\section{Analysis of the Task Completion Inference Module}
\label{sec:appendix-success-inference-module}
\begin{figure}[ht]
  \centering
  \includegraphics[width=0.85\linewidth]{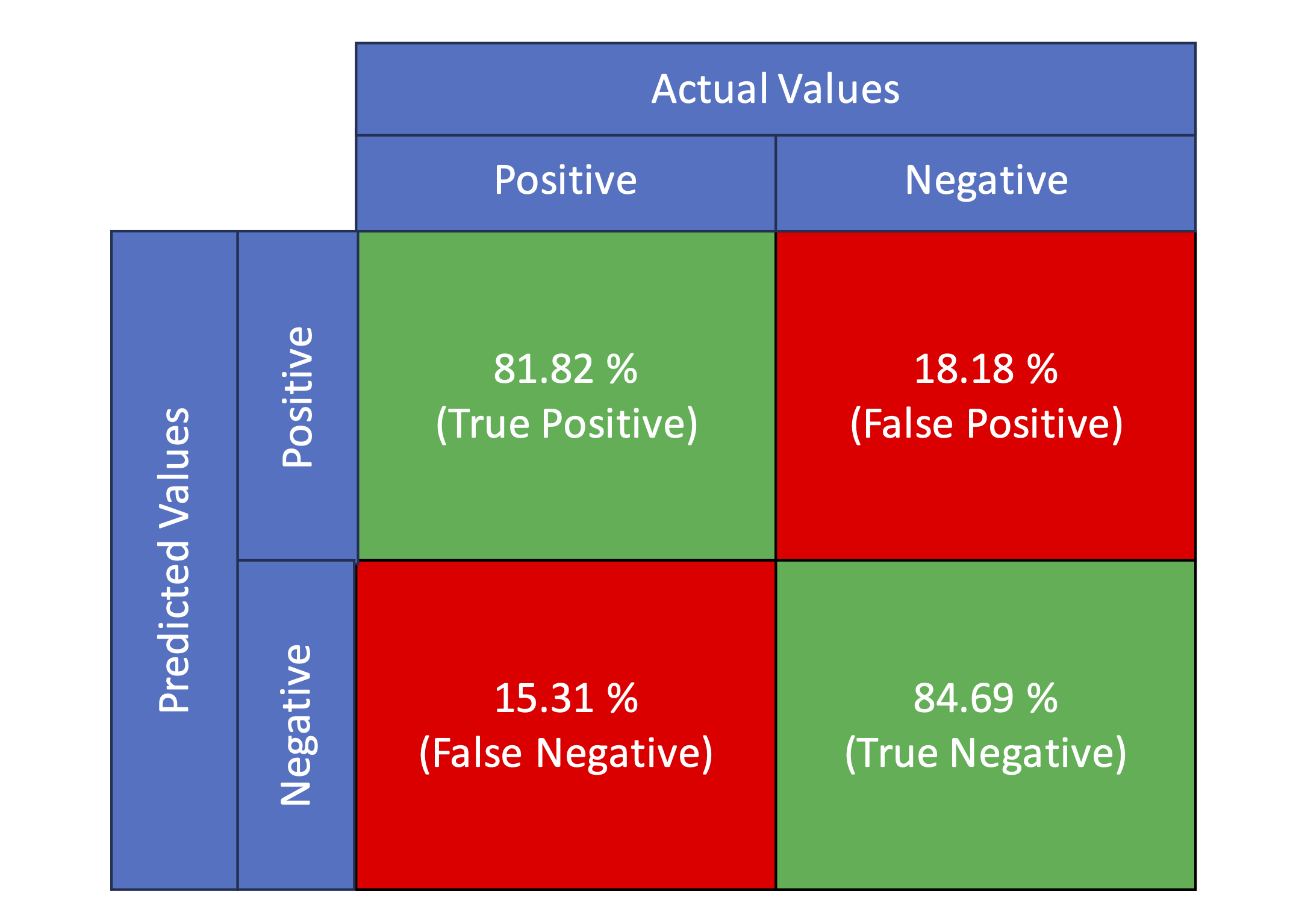}
  \caption{Confusion matrix highlighting the capabilities of our task completion inference module. We observe a true positive rate of 81.82 \% and a true negative rate rate of 84.69 \%.}
  \vspace{-1mm}
  \label{fig:confusion-matrix}
\end{figure}

Given a task, a sequence of actions are performed by the agent which in turn constructs the final scene graph that reflects the up-to-date states of the objects in the scene and their relationships. 
The reasoning capability of the GPT-4 is leveraged to determine the completion of the task after a scene graph is incorporated into a prompt. 
To measure the performance of the task completion inference module , the trajectories are collected and frames of the trajectories are analyzed through human supervision to check whether the task has been completed or not, and whether it matches the result obtained from the task completion module.
In Tab.~\ref{tab:false-positives} we observe the false positives obtained when the task completion module labels the task as successfully completed but the trajectory is incomplete or wrong.
For our experiment , we select 20 tasks and collect their trajectories after which we manually infer the accuracy of the task completion inference module for the trajectories which have been marked successful.
From Tab.~\ref{tab:false-positives}, it is observed that we get approximately 16 errors in 100 successful trajectories collected.
In addition, we also note that for the task where the primitive action is 'Push', the performance of our task completion inference module will be poor along with cases of long horizon tasks which include color and spatial awareness.
We also choose another set of tasks and record both the trajectories labelled as successful by the inference module as well as the trajectories labelled as unsuccessful.
A confusion matrix highlighting the true positives, false positives, false negatives and true negatives based on the result provided by the task completion inference module is shown in Fig.~\ref{fig:confusion-matrix}

\begin{table}[h]
    \centering
    \scriptsize
        \captionof{table}{False positives observed through manual checking of the trajectories collected from 20 tasks .}
    \begin{tabular}{lcc}
      \toprule
     Tasks & Errors & Trajectories \\
      \midrule
      Close the drawer & 1 & 200  \\
      Move the green block into open drawer & 14 & 200 \\
      Align the red block and the stick along the x-axis & 8 & 200\\
      Gather the red stick and red block into the green bin  & 22 & 200\\
      Arrange all the red blocks in a straight line  & 29 & 200 \\
      Stack all the red blocks on top of each other  & 0 & 200 \\
      Pick up the red block  & 1 & 200 \\
      Swap the positions of the red block and turbo airplane toy & 21 & 225 \\
      Arrange all the red blocks in a straight line  & 29 & 200 \\
      Move the green block into the bowl  & 28 & 180 \\
      Push the green block towards the left  & 169 & 200\\
      Create a tower by stacking one block of each color & 1 & 93 \\
      on top of each other & & \\
      Move all the blocks to the corresponding colored stickers & 69 & 200\\
      Move the blue block and the red block on opposite ends & 53 & 200\\
      of the blue line& & \\
      Place the blue block and the red block side by side  & 35 & 200 \\
      on the blue line & & \\
      Press the catapult button to trigger the catapult & 0 & 100 \\
      Launch the green block using the catapult & 18 & 200 \\
      Sort all the blocks by color & 14 & 93 \\
      Move the turbo airplane toy to the front and the & 15 & 200 \\
      red block to the back & & \\
      Move all the red blocks to the left of the green blocks  & 6 &206 \\\cline{1-2}

      \toprule
      Total  &  \textbf{603} & 3697\\ 
      \bottomrule
    \end{tabular}
    \vspace{-2mm}
    \label{tab:false-positives}
\end{table}

\section{Prompts for Large Language Model}
\label{sec:prompts for large language model}
The complete prompts consist of base prompts, which include instructions along with a few shots of examples, as well as appended input data. Listing~\ref{listing:prompt-construction} illustrates the construction of prompts for task proposal, task decomposition and success inference in the form of Python code.
We also present the base prompts for task proposal (Listing~\ref{listing:proposal}), task decomposition (Listing~\ref{listing:decomposition}) and success inference (Listing~\ref{listing:inference}). A small number of examples are provided solely to help the LLM understand its task and the desired output format.

\begin{lstlisting}[caption=Prompt construction., label=listing:prompt-construction, frame=none, basicstyle=\small\ttfamily, commentstyle=\color{citecolor}\small\ttfamily, breaklines=true, escapeinside={(*}{*)}]
task_proposal_prompt = task_proposal_base_prompt + '\n' + '```' + '\n' + 'scene graph:' + '\n' + scene_graph + 'tasks:'
task_decomposition_prompt = task_decomposition_base_prompt + '\n' + '```' + '\n' + 'task description: ' + task_desc + '\n' + 'scene graph:' + '\n' + scene_graph + 'reasoning: '
success_inference_prompt = success_inference_base_prompt + '\n' + '```' + '\n' + 'task description: ' + task_desc + '\n' + 'scene graph list:' + '\n' + scene_graph_list_str + 'success metric: '
\end{lstlisting}

ht
\clearpage
\onecolumn
\begin{lstlisting}[caption=\textbf{Base prompts for task proposal.} Instructions and output format are provided in the base prompts for task proposal., label=listing:proposal, backgroundcolor=\color{lightbrown}, breaklines=true, frame=none, basicstyle=\small\ttfamily, commentstyle=\color{citecolor}\small\ttfamily,columns=fullflexible, escapeinside={(*}{*)}]
You are a curious baby. Given a scene graph, the goal is to propose as many diverse tasks as possible for a robot arm with a gripper. The nodes in the scene graph indicate the name, the state, the position and the bounding box (in the unit of meter) of an object. The positive direction of the x-axis represents the front, and the negative direction represents the rear. The positive direction of the y-axis represents the right side, and the negative direction represents the left side. The positive direction of the z-axis represents upward, and the negative direction represents downward. The edges indicate the spatial relationships between the objects, and no edges between two objects (nodes) means the two objects are far apart. The position of the robot is (0.0, 0.0, 0.0). 
Note: (1) The proposed tasks should be as diverse as possible; (2) It is necessary to consider the objects present in the scene, their state, attributes, position and spatial relationships; (3) The proposed tasks are unrelated to each other; (4) One type of object may have multiple instances, nodes use numbers to distinguish them. However, when proposing a task, assume that you don't know the numerical labels corresponding to the instances and you only know the quantity of instances. 
More importantly, you should consider the primitive actions the robot arm could take: 
Pick(obj_name), PlaceOn(obj_name), PlaceAt(place_pos), Push(obj_name, direction, distance), RevoluteJointOpen(obj_name), RevoluteJointClose(obj_name), PrismaticJointOpen(obj_name), PrismaticJointClose(obj_name), Press(obj_name),
and then propose feasible tasks.
Here I will show you the implementation of the primitive actions for your reference.
Pick(obj_name): Approach the object, close the gripper to grasp it and lift it up (Parameters: obj_name -- the name of the object which would be picked). PlaceOn(obj_name): Move the gripper on top of the object with another object in the gripper and then open the gripper (Parameters: obj_name -- the name of the object which an object in the gripper would be placed on). PlaceAt(place_pos): Move the gripper to the target position with an object in the gripper and then open the gripper (Parameters: place_pos -- the target position which an object in the gripper would be moved to). Push(obj_name, direction, distance): Close the gripper and then push the object in the specified direction by a specified distance (Parameters: obj_name -- the name of the object which would be pushed; direction -- the direction which the object would be moved in, it is direction vector [x, y]; distance -- the distance which the object would be moved by, the distance is in the unit of meter). PrismaticJointOpen(obj_name): Open the object with a prismatic joint (Parameters: obj_name -- the name of the handle of the object with a prismatic joint). PrismaticJointClose(obj_name): Close the object with a prismatic joint (Parameters: obj_name -- the name of the handle of the object with a prismatic joint). Press(obj_name): Close the gripper and then press the object (Parameters: obj_name -- the name of the object which should be pressed).
Below is an example to show the format:
```
scene graph:
  [Nodes]:
    - object_A (state) -- position: (x, y, z), x_range: [min, max], y_range: [min, max], z_range: [min, max]
    - object_B (state) -- position: (x, y, z), x_range: [min, max], y_range: [min, max], z_range: [min, max]
    - object_B 1 (state) -- position: (x, y, z), x_range: [min, max], y_range: [min, max], z_range: [min, max]
    - ...
  [Edges]:
    object_B 1 -> relationship -> object_A
tasks:
 - task description 1
 - task description 2
 - ...
```
Now you should read the scene graph I provide with you at first, and then think about diverse and feasible tasks. Think step by step, and imagine the process to accomplish the task with the primitive actions provided with you: Pick(obj_name), PlaceOn(obj_name), PlaceAt(place_pos), Push(obj_name, direction, distance), RevoluteJointOpen(obj_name), RevoluteJointClose(obj_name), PrismaticJointOpen(obj_name), PrismaticJointClose(obj_name), Press(obj_name).
REMEMBER, YOU MUST make sure that you don't know the numerical labels corresponding to the instances, you only know the quantity of instances; DO NOT use primitive actions explicitly in the task description. 
\end{lstlisting}

\begin{lstlisting}[caption=\textbf{Base prompts for task decomposition.} The base prompts include instructions and a small number of examples which are provided solely to help the LLM understand its task and clarify the desired output format., label=listing:decomposition, backgroundcolor=\color{lightbrown}, frame=none, basicstyle=\small\ttfamily, commentstyle=\color{citecolor}\small\ttfamily,columns=fullflexible, breaklines=true, escapeinside={(*}{*)}]
You are a robot with a single arm in a tabletop robot manipulation environment. 
Given a task description and a scene graph, the goal is to decompose the task into subtasks and to call corresponding primitive actions, which, when performed in sequence, would solve the input task. The nodes in the scene graph indicate the name, the state, the position and the bounding box (in the unit of meter) of an object. The positive direction of the x-axis represents the front, and the negative direction represents the rear. The positive direction of the y-axis represents the right side, and the negative direction represents the left side. The positive direction of the z-axis represents upward, and the negative direction represents downward. One type of object may have multiple instances, nodes use numbers to distinguish them. The edges indicate the spatial relationships between the objects, and no edges between two objects (nodes) means the two objects are far apart. The position of the robot is (0.0, 0.0, 0.0). 
You should consider the primitive actions the robot arm could take: 
Pick(obj_name), PlaceOn(obj_name), PlaceAt(place_pos), Push(obj_name, direction, distance), RevoluteJointOpen(obj_name), RevoluteJointClose(obj_name), PrismaticJointOpen(obj_name), PrismaticJointClose(obj_name), Press(obj_name), and decompose a task into feasible sub-tasks. 
Here I will show you the implementation of the primitive actions for your reference.
Pick(obj_name): Approach the object, close the gripper to grasp it and lift it up (Parameters: obj_name -- the name of the object which would be picked). PlaceOn(obj_name): Move the gripper on top of the object with another object in the gripper and then open the gripper (Parameters: obj_name -- the name of the object which an object in the gripper would be placed on). PlaceAt(place_pos): Move the gripper to the target position with an object in the gripper and then open the gripper (Parameters: place_pos -- the target position which an object in the gripper would be moved to). Push(obj_name, direction, distance): Close the gripper and then push the object in the specified direction by a specified distance (Parameters: obj_name -- the name of the object which would be pushed; direction -- the direction which the object would be moved in, it is direction vector [x, y]; distance -- the distance which the object would be moved by, the distance is in the unit of meter). PrismaticJointOpen(obj_name): Open the object with a prismatic joint (Parameters: obj_name -- the name of the handle of the object with a prismatic joint). PrismaticJointClose(obj_name): Close the object with a prismatic joint (Parameters: obj_name -- the name of the handle of the object with a prismatic joint). Press(obj_name): Close the gripper and then press the object (Parameters: obj_name -- the name of the object which should be pressed).
Below are some examples:
```
task description: move the red block onto the plate, the blue block onto the red block, and the green block on the blue block
scene graph:
  [Nodes]:
    - red block -- position: [0.40, -0.20, 0.08], x_range: [0.37, 0.43], y_range: [-0.23, -0.17], z_range: [0.05, 0.11]
    - blue block -- position: [0.30, -0.20, 0.08], x_range: [0.27, 0.33], y_range: [-0.30, -0.1], z_range: [0.05, 0.12]
    - green block -- position: [0.05, -0.26, 0.08], x_range: [0.02, 0.08], y_range: [-0.29, -0.23], z_range: [0.05, 0.11]
    - plate -- position: [0.52, -0.21, 0.09], x_range: [0.44, 0.60], y_range: [-0.30, -0.13], z_range: [0.07, 0.11]
  [Edges]:
reasoning: Objects should be stacked from bottom to top. Firstly, move the red block onto the plate. Secondly, move the blue block onto the red block. Thirdly, move the green block onto the blue block. "move onto" can be done via Pick and PlaceOn.
answer:
 - 1. move the red block onto the plate | [Pick('red block'); PlaceOn('plate)]
 - 2. move the blue block onto the red block | [Pick('blue block'); PlaceOn('red block')]
 - 3. move the green block onto the blue block | [Pick('green block'); PlaceOn('blue block')]
```
```
task description: move the block at the farthest left to the right
scene graph:
  [Nodes]:
    - red block -- position: [0.40, -0.20, 0.08], x_range: [0.37, 0.43], y_range: [-0.23, -0.17], z_range: [0.05, 0.11]
    - blue block -- position: [0.30, -0.20, 0.08], x_range: [0.27, 0.33], y_range: [-0.30, -0.1], z_range: [0.05, 0.12]
    - green block -- position: [0.05, -0.26, 0.08], x_range: [0.02, 0.08], y_range: [-0.29, -0.23], z_range: [0.05, 0.11]
    - plate -- position: [0.52, -0.21, 0.09], x_range: [0.44, 0.60], y_range: [-0.30, -0.13], z_range: [0.07, 0.11]
  [Edges]:
reasoning: "farthest left" refers to the y-coordinate with the smallest value, and the y of the green block which is -0.26 is the smallest, so the green block is at the farthest left. so the robot should move the green block. Since the positive direction of the y-axis represents the right side, a potential right position is [0.05, 0.33, 0.08], so the robot should place the green block at [0.05, 0.33, 0.08].
answer:
 - 1. move the green block at [0.05, 0.33, 0.08] | [Pick('green block'); PlaceAt([0.05, 0.33, 0.08])]
```
```
task description: move the red block in the drawer 
scene graph:
  [Nodes]:
    - red block -- position: [0.40, -0.20, 0.08], x_range: [0.37, 0.43], y_range: [-0.23, -0.17], z_range: [0.05, 0.11]
    - drawer (closed) -- position: [0.20, -0.40, 0.18], x_range: [0.05, 0.35], y_range: [-0.55, -0.25], z_range: [0.05, 0.31]
    - drawer handle -- position: [0.25, -0.35, 0.18], x_range: [0.22, 0.28], y_range: [-0.38, -0.32], z_range: [0.17, 0.19]
  [Edges]:
reasoning: the drawer starts off closed. It needs to be opened before objects can be moved into it. After the task is done, it needs to be closed. So firstly, open the drawer. Secondly, move the red block into the table. Thirdly, close the drawer. 
answer:
 - 1. open the drawer | [PrismaticJointOpen('drawer handle')]
 - 2. move the red block into the table | [Pick('red block'); PlaceOn('drawer')]
 - 3. close the drawer | [PrismaticJointClose('drawer handle')]
```
```
task description: move the bowl into the microwave
scene graph:
  [Nodes]:
    - microwave (closed) -- position: [0.20, -0.40, 0.18], x_range: [0.05, 0.35], y_range: [-0.55, -0.25], z_range: [0.05, 0.31]
    - microwave handle -- position: [0.32, -0.40, 0.18], x_range: [0.30, 0.34], y_range: [-0.42, -0.37], z_range: [0.17, 0.19]
    - microwave button -- position: [0.32, -0.40, 0.08], x_range: [0.30, 0.34], y_range: [-0.42, -0.37], z_range: [0.06, 0.10]
    - bowl -- position: [0.40, -0.20, 0.08], x_range: [0.37, 0.43], y_range: [-0.23, -0.17], z_range: [0.05, 0.11]
  [Edges]:
reasoning: The microwave starts off closed. It needs to be opened before objects can be put in them. After the task is done, it needs to be closed. Microwaves should be opened by pressing a button. So firstly, open the door of the microwave. Secondly, move the bowl into the microwave. Thirdly, close the door of the microwave.
answer:
 - 1. open the door of the microwave | [Press('microwave button')]
 - 2. move the bowl into the microwave | [Pick('bowl'); PlaceOn('microwave')]
 - 3. close the door of the microwave | [RevoluteJointClose('microwave handle')]
```
```
task description: move the red block to the left of the plate
scene graph:
  [Nodes]:
    - red block -- position: [0.40, -0.20, 0.08], x_range: [0.37, 0.43], y_range: [-0.23, -0.17], z_range: [0.05, 0.11]
    - green block -- position: [0.05, -0.26, 0.08], x_range: [0.02, 0.08], y_range: [-0.29, -0.23], z_range: [0.05, 0.11]
    - plate -- position: [0.52, -0.21, 0.09], x_range: [0.44, 0.60], y_range: [-0.30, -0.13], z_range: [0.07, 0.11]
  [Edges]:
reasoning: The positive direction of the y-axis represents the right side, and the negative direction represents the left side. The x of the plate is 0.52. So the x which the red block would be placed at: x = 0.52. The y_range of the plate is [-0.30, -0.13]. So the y which the red block would be placed at: y = -0.30 - about 0.06 = -0.36. The original z of the red block is 0.08. So the z which the red block would be placed at: z = 0.08 + about 0.02. So move the red block at [0.52, -0.36, 0.10].
answer:
 - 1. move the red block at [0.52, -0.36, 0.10] | [Pick('red block'); PlaceAt([0.52, -0.36, 0.10])]
```

Now I'd like you to help me decompose the following task into subtasks and call corresponding primitive actions. You should read the task description and the scene graph I provide with you, and think about how to decompose the task. Think step by step, and imagine the process to accomplish the task with the primitive actions provided with you: Pick(obj_name), PlaceOn(obj_name), PlaceAt(place_pos), Push(obj_name, direction, distance), RevoluteJointOpen(obj_name), RevoluteJointClose(obj_name), PrismaticJointOpen(obj_name), PrismaticJointClose(obj_name), Press(obj_name). Use PrismaticJointOpen and PrismaticJointClose to open and close something that has a prismatic joint, like drawer. Note that the second and third parameters of Push are list of float and float respectively, do not use expressions composed of variables. 
\end{lstlisting}

\begin{lstlisting}[caption=\textbf{Base prompts for success inference.} The base prompts include instructions and a limited number of examples intended solely to enable the large language model to understand its task and output format and to engage in thorough reasoning before providing answers., label=listing:inference, backgroundcolor=\color{lightbrown}, frame=none, basicstyle=\small\ttfamily, commentstyle=\color{citecolor}\small\ttfamily,columns=fullflexible, breaklines=true, escapeinside={(*}{*)}]
You are a robot with a single arm in a tabletop robot manipulation environment. 
Given a task description and a list of scene graphs, the goal is to infer if the task has been completed successfully. The list of scene graphs are arranged in chronological order (the first is the initial scene graph, and the last is the scene graph after the policy is executed). The nodes in the scene graph indicate the name, the state, the position and the bounding box (in the unit of meter) of an object. The positive direction of the x-axis represents the front, and the negative direction represents the rear. The positive direction of the y-axis represents the right side, and the negative direction represents the left side. The positive direction of the z-axis represents upward, and the negative direction represents downward. One type of object may have multiple instances, nodes use numbers to distinguish them. The edges indicate the spatial relationships between the objects, and no edges between two objects (nodes) means the two objects are far apart. The position of the robot is (0.0, 0.0, 0.0). 
Note that you should firstly reason whether the task is completed based on the task description and scene graphs, then output the answer. The answer should be "yes" or "no" or "not sure". Below are some examples:
```
task description: move the blue block on the plate
scene graph list:
  ----------
  [Nodes]:
    - red block -- position: [0.40, -0.20, 0.08], x_range: [0.37, 0.43], y_range: [-0.23, -0.17], z_range: [0.05, 0.11]
    - blue block -- position: [0.30, -0.20, 0.08], x_range: [0.27, 0.33], y_range: [-0.23, -0.17], z_range: [0.05, 0.12]
    - plate -- position: [0.52, -0.21, 0.09], x_range: [0.44, 0.60], y_range: [-0.30, -0.13], z_range: [0.07, 0.11]
  [Edges]:
  ----------
  [Nodes]:
    - red block -- position: [0.48, -0.20, 0.15], x_range: [0.45, 0.51], y_range: [-0.23, -0.17], z_range: [0.13, 0.18]
    - blue block -- position: [0.30, -0.20, 0.08], x_range: [0.27, 0.33], y_range: [-0.23, -0.17], z_range: [0.05, 0.12]
    - plate -- position: [0.52, -0.21, 0.09], x_range: [0.44, 0.60], y_range: [-0.30, -0.13], z_range: [0.07, 0.11]
  [Edges]:
    - red block -> on top of -> blue block
success metric: The blue block is on the plate
reasoning: The scene graph indicates the blue block is not on the plate. So the task was not accomplished.
answer:
  no
```
```
task description: stack the red block and the blue block on the plate
scene graph list:
  ----------
  [Nodes]:
    - red block -- position: [0.40, -0.29, 0.08], x_range: [0.37, 0.43], y_range: [-0.32, -0.26], z_range: [0.05, 0.11]
    - blue block -- position: [0.32, -0.20, 0.08], x_range: [0.29, 0.35], y_range: [-0.23, -0.17], z_range: [0.05, 0.11]
    - plate -- position: [0.52, -0.21, 0.09], x_range: [0.44, 0.60], y_range: [-0.30, -0.13], z_range: [0.07, 0.11]
  [Edges]:
  ----------
  [Nodes]:
    - red block -- position: [0.40, -0.29, 0.08], x_range: [0.37, 0.43], y_range: [-0.32, -0.26], z_range: [0.05, 0.11]
    - blue block -- position: [0.48, -0.20, 0.15], x_range: [0.45, 0.51], y_range: [-0.23, -0.17], z_range: [0.12, 0.18]
    - plate -- position: [0.52, -0.21, 0.09], x_range: [0.44, 0.60], y_range: [-0.30, -0.13], z_range: [0.07, 0.11]
  [Edges]:
    - blue block -> on top of -> plate
  ----------
  [Nodes]:
    - red block -- position: [0.48, -0.20, 0.21], x_range: [0.45, 0.51], y_range: [-0.23, -0.17], z_range: [0.18, 0.24]
    - blue block -- position: [0.48, -0.20, 0.15], x_range: [0.45, 0.51], y_range: [-0.23, -0.17], z_range: [0.12, 0.18]
    - plate -- position: [0.52, -0.21, 0.09], x_range: [0.44, 0.60], y_range: [-0.30, -0.13], z_range: [0.07, 0.11]
  [Edges]:
    - blue block -> on top of -> plate
    - red block -> on top of -> blue block
success metric: The blue block is on top of the plate and the red block is on top of the blue block.
reasoning: The scene graph indicates the blue block is on top of the plate and the red block is on top of the blue block. So "stack the red block and the blue block on the plate" has been accomplished.
answer:
  yes
```
```
task description: move the red block to the left of the plate
scene graph list:
  ----------
  [Nodes]:
    - red block -- position: [0.40, -0.20, 0.08], x_range: [0.37, 0.43], y_range: [-0.23, -0.17], z_range: [0.05, 0.11]
    - green block -- position: [0.05, -0.26, 0.08], x_range: [0.02, 0.08], y_range: [-0.29, -0.23], z_range: [0.05, 0.11]
    - plate -- position: [0.52, -0.21, 0.09], x_range: [0.44, 0.60], y_range: [-0.30, -0.13], z_range: [0.07, 0.11]
  [Edges]:
  ----------
  [Nodes]:
    - red block -- position: [0.40, -0.05, 0.08], x_range: [0.37, 0.43], y_range: [-0.08, -0.02], z_range: [0.05, 0.11]
    - green block -- position: [0.05, -0.26, 0.08], x_range: [0.02, 0.08], y_range: [-0.29, -0.23], z_range: [0.05, 0.11]
    - plate -- position: [0.52, -0.21, 0.09], x_range: [0.44, 0.60], y_range: [-0.30, -0.13], z_range: [0.07, 0.11]
  [Edges]:
success metric: The red block is on the left of the plate. In other words, the y of the red block is less than the minimum value of the plate's y_range.
reasoning: The y of the red block is -0.05, which is larger than the minimum value of the plate's y_range (-0.30). So "move the red block to the left of the plate" has been accomplished.
answer:
  no
```
Now I'd like you to help me infer whether the proposed task is completed. You should read the task description and the scene graph list I provide with you, and then complete the "success metric", "reasoning" and "answer".
\end{lstlisting}

\newpage
\section{Additional visualizations}
We also present additional images from our project which shows the robotic arm in the tabletop environment performing various different tasks. The images are from the trajectories of the 60 tasks on which the policy is trained on. In addition, there are also images from the zero-shot generalization and few-shot generalization experiments. The full trajectories (including the animation) can also be found on the project website \href{https://bbsea-embodied-ai.github.io}{https://bbsea-embodied-ai.github.io}. 

\begin{figure*}[ht]
  \centering
  \includegraphics[width=0.85\linewidth]{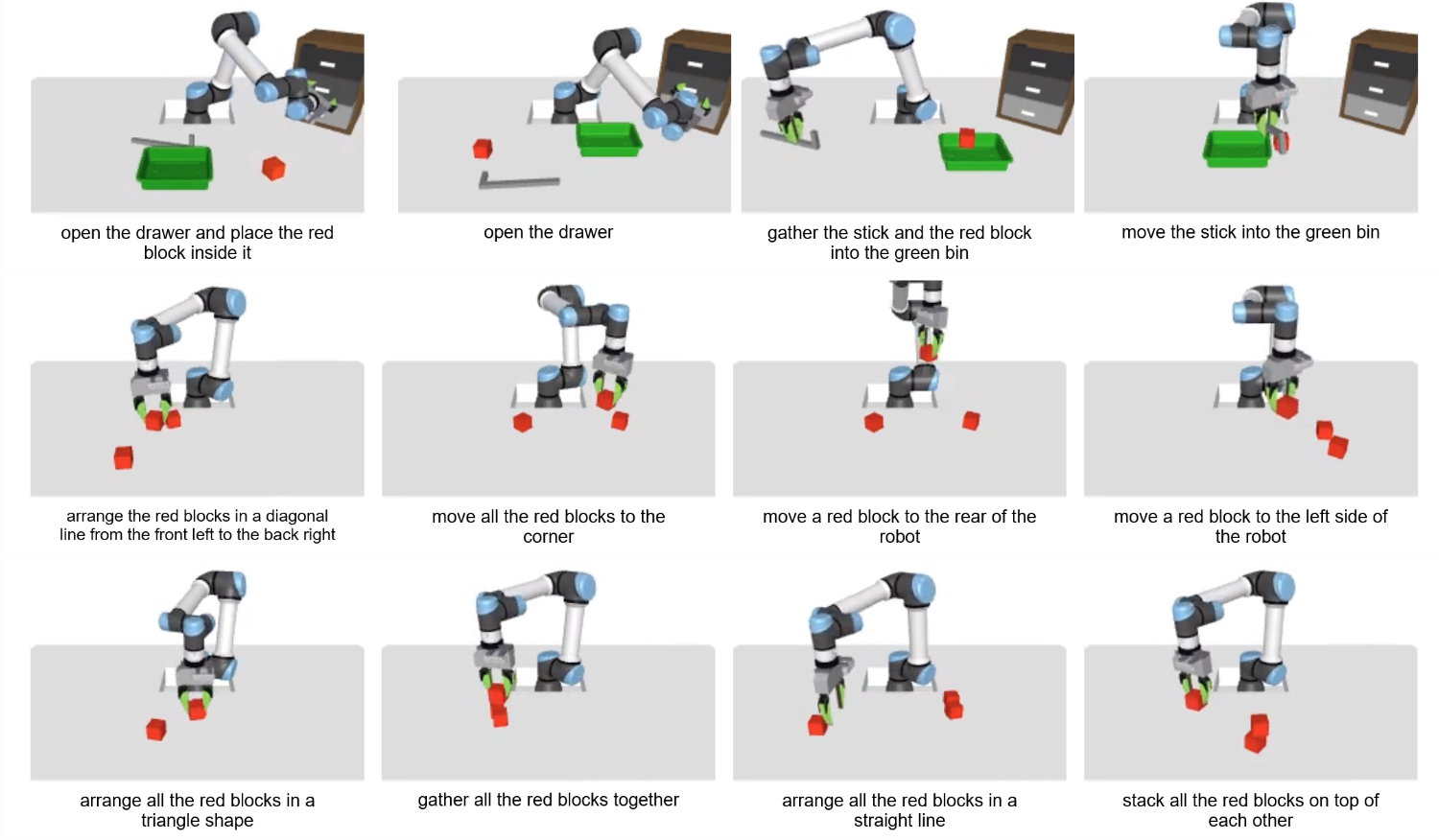}
  \vspace{-1mm}
  \label{fig:vis-1}
\end{figure*}

\begin{figure*}[ht]
  \centering
  \includegraphics[width=0.85\linewidth]{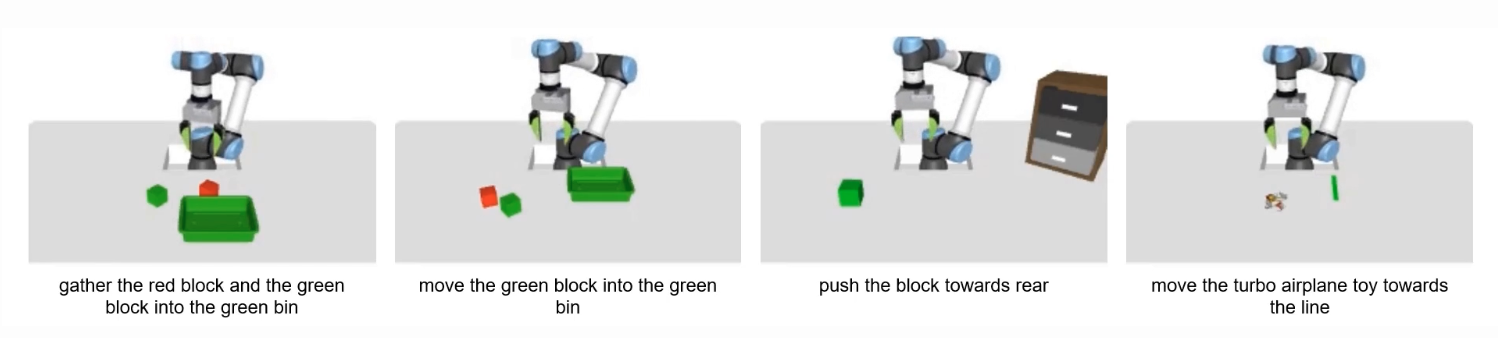}
  \vspace{-1mm}
  \label{fig:vis-2}
\end{figure*}

\begin{figure*}[t]
  \centering
  \includegraphics[width=0.85\linewidth]{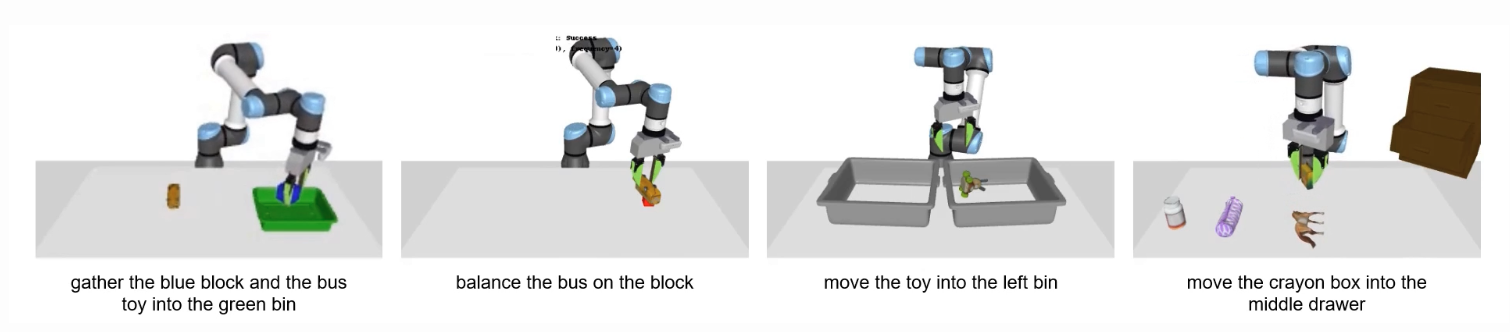}
  \vspace{-1mm}
  \label{fig:vis-3}
\end{figure*}

\begin{figure*}[t]
  \centering
  \includegraphics[width=0.85\linewidth]{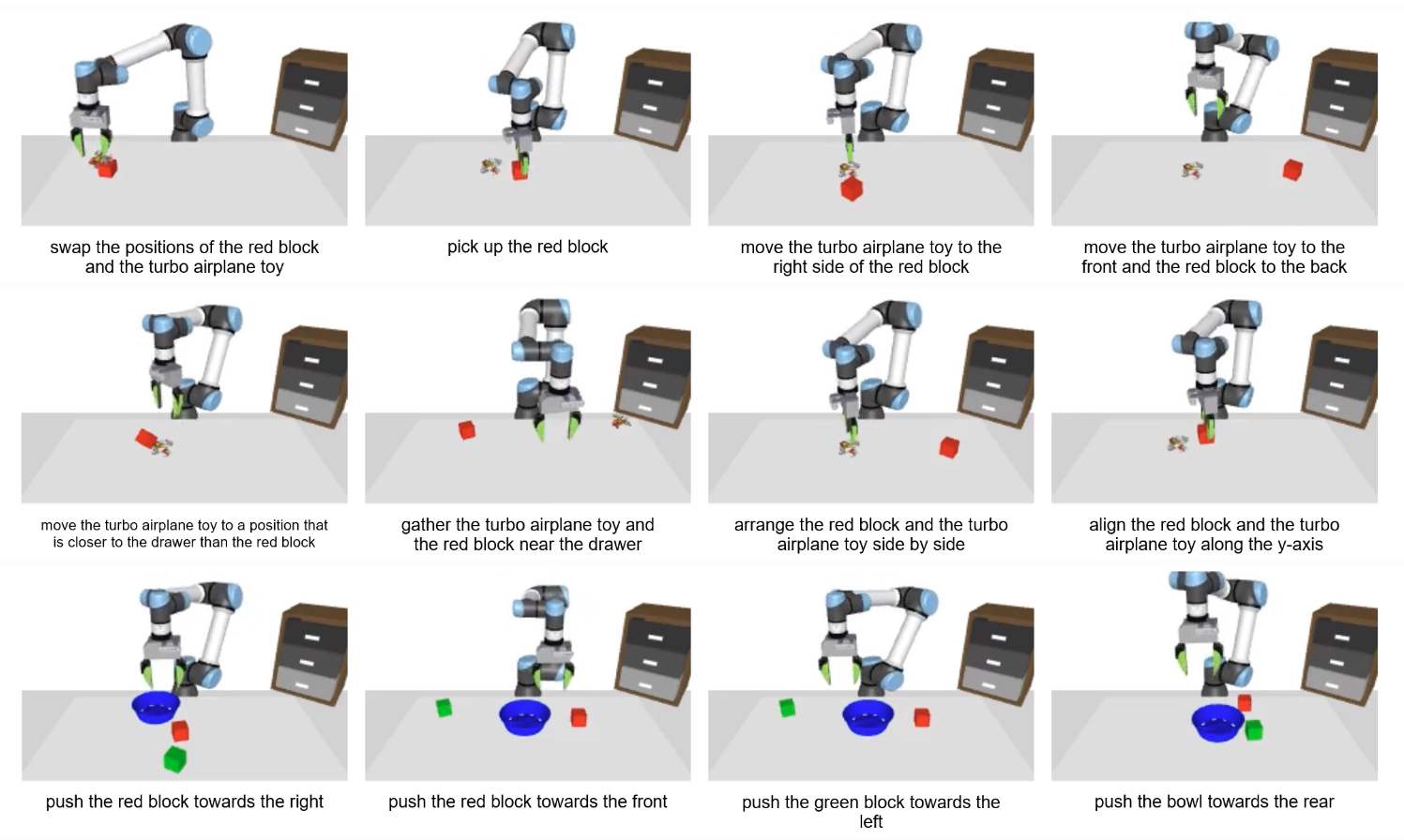}
  \vspace{-1mm}
  \label{fig:vis-4}
\end{figure*}

\begin{figure*}[t]
  \centering
  \vspace{-6mm}
  \includegraphics[width=0.85\linewidth]{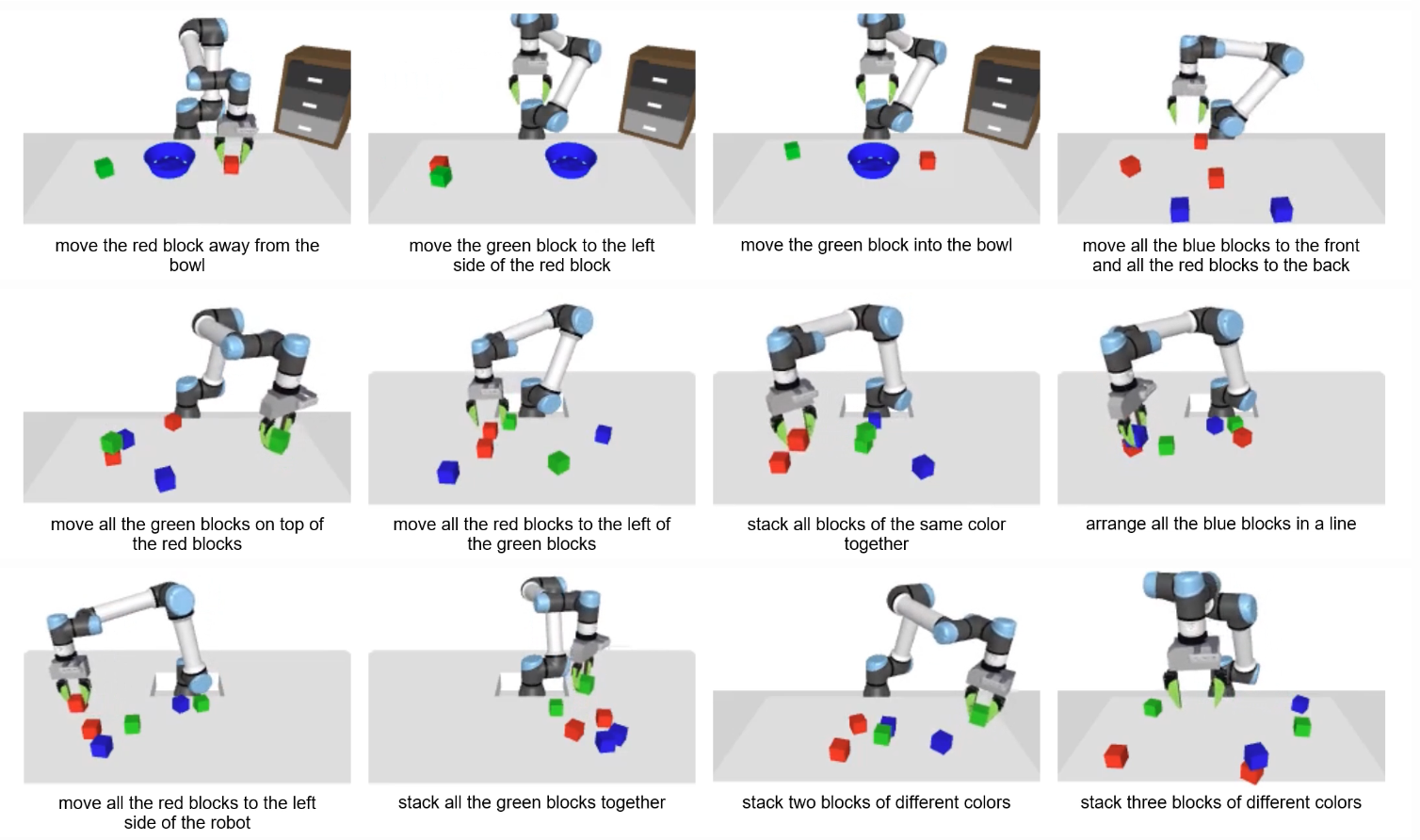}
  \vspace{-1mm}
  \label{fig:vis-5}
\end{figure*}

\begin{figure*}[t]
  \centering
  \includegraphics[width=0.85\linewidth]{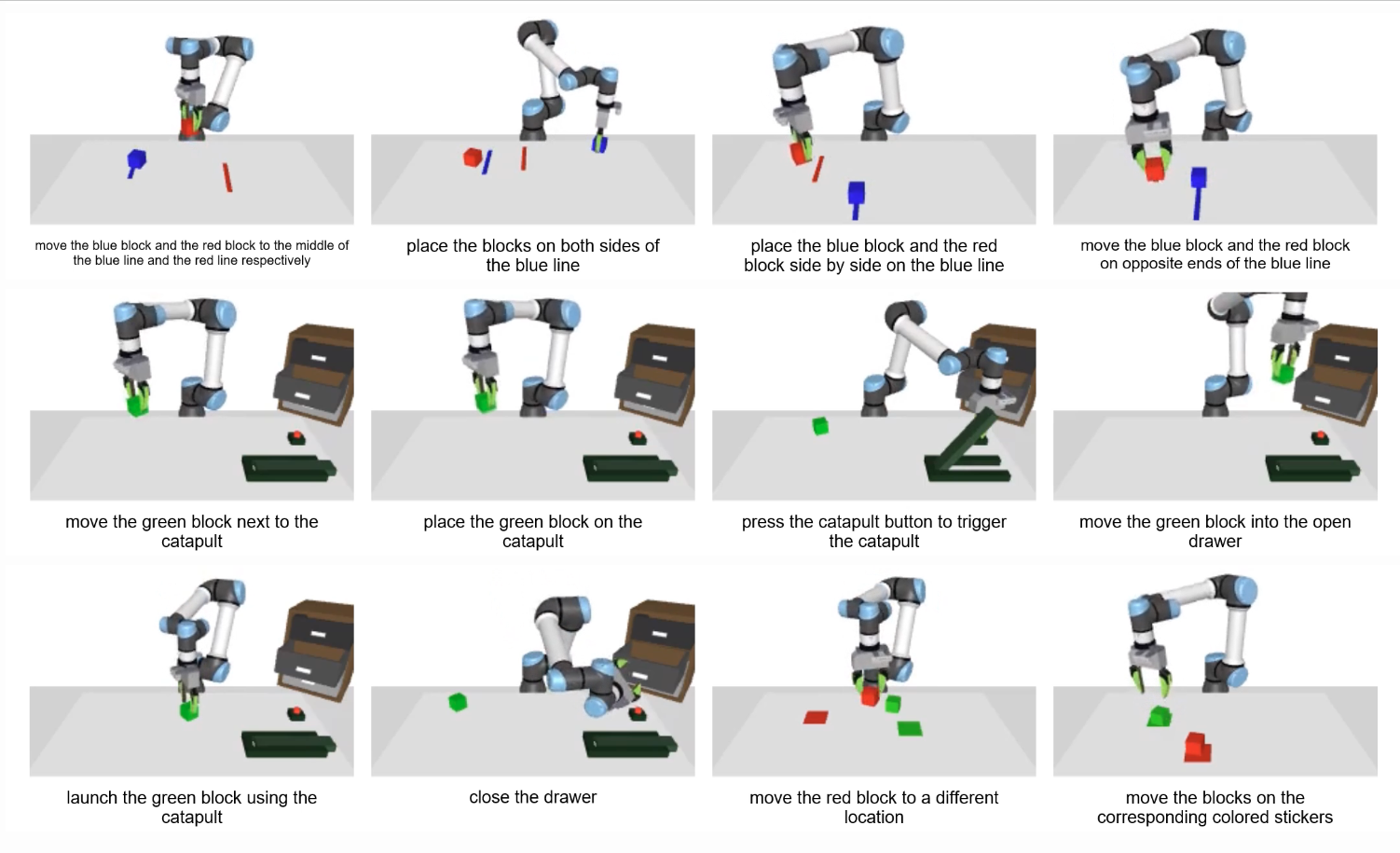}
  \vspace{-1mm}
  \label{fig:vis-6}
\end{figure*}

\twocolumn